\documentclass[letterpaper, 10 pt, conference]{ieeeconf}  

\IEEEoverridecommandlockouts                              

\usepackage[utf8]{inputenc}
\usepackage[english]{babel}
\usepackage[noadjust]{cite} 

\usepackage{graphicx} 
\usepackage{comment}
\usepackage[noend]{algpseudocode} 
\usepackage[ruled,linesnumbered]{algorithm2e} 
\usepackage{amsmath}
\usepackage{amssymb} 
\usepackage{mathrsfs} 
\usepackage{fdsymbol} 
\usepackage{mathtools} 
\usepackage{multirow}
\usepackage{indentfirst}
\usepackage{booktabs}
\usepackage{tabularx}
\usepackage{subcaption}
\usepackage{color}
\usepackage{hyperref}
\usepackage{makecell}
\usepackage{enumerate}

\usepackage{amsthm}

\newtheorem{theorem}{Theorem}

\newtheorem{lemma}{Lemma}

\theoremstyle{definition}

\newcommand{\RN}[1]{%
  \textup{\uppercase\expandafter{\romannumeral#1}}%
}

\title{\LARGE \bf
Decentralized Deadlock-free Trajectory Planning for Quadrotor Swarm in Obstacle-rich Environments - Extended version
}

\author{Jungwon Park, Inkyu Jang, and H. Jin Kim$^{1}$
\thanks{$^{1}$The authors are with the Department of Mechanical and Aerospace Engineering, Seoul National University (SNU), and Automation and Systems Research Institute (ASRI), Seoul 08826, South Korea
        {\tt\footnotesize \{qwerty35, leplusbon, hjinkim\}@snu.ac.kr}}%
}

\begin{document}

\maketitle
\thispagestyle{empty}
\pagestyle{empty}

\begin{abstract}
This paper presents a decentralized multi-agent trajectory planning (MATP) algorithm that guarantees to generate a safe, deadlock-free trajectory in an obstacle-rich environment under a limited communication range.
The proposed algorithm utilizes a grid-based multi-agent path planning (MAPP) algorithm for deadlock resolution, and we introduce the subgoal optimization method to make the agent converge to the waypoint generated from the MAPP without deadlock.
In addition, the proposed algorithm ensures the feasibility of the optimization problem and collision avoidance by adopting a linear safe corridor (LSC).
We verify that the proposed algorithm does not cause a deadlock in both random forests and dense mazes regardless of communication range, and it outperforms our previous work in flight time and distance.
We validate the proposed algorithm through a hardware demonstration with ten quadrotors.
\end{abstract}

\section{INTRODUCTION}
\label{introduction}
Multi-agent trajectory planning (MATP) is essential to utilize a large group of unmanned vehicles in various applications such as search and rescue, surveillance, and transportation.
Among many MATP algorithms, decentralized approaches have received much attention due to their high scalability and low computation load, which enables online planning. However, most decentralized algorithms do not consider obstacles \cite{zhou2017fast,luis2020online,chen2022recursive}
or have a risk of causing a deadlock in an obstacle-rich environment \cite{zhou2021ego,tordesillas2021mader,toumieh2022decentralized}.

\begin{figure}[t]
\centering
\includegraphics[width = 0.75\linewidth]{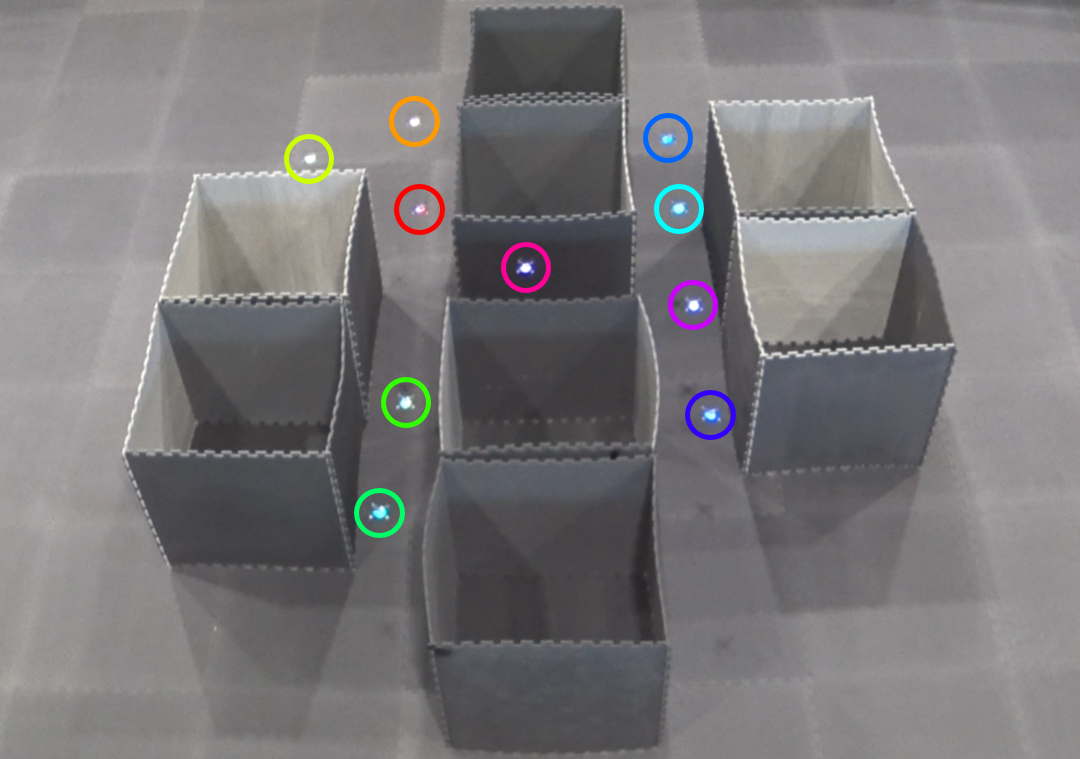}
\caption{
Experiment with 10 quadrotors in a dense maze.
}
\label{fig: thumbnail}
\vspace{-5mm}
\end{figure}

This paper presents a decentralized multi-agent trajectory planning (MATP) algorithm that guarantees to generate a safe, deadlock-free trajectory in a cluttered environment.
The proposed method solves a deadlock through the following three steps. First, we compute the waypoint of each agent using a decentralized grid-based multi-agent path planning (MAPP) algorithm. 
Then, we optimize a subgoal of each agent considering the collision constraints and communication range so that the agent can reach the waypoint without deadlock. Finally, we conduct trajectory optimization to make the agent converge to the waypoint. As a result, the proposed algorithm allows the agent to reach the goal by following the waypoints from the grid-based MAPP.
We utilize a linear safe corridor (LSC) \cite{park2022online} to guarantee the feasibility of the optimization problem and collision avoidance.
Moreover, the proposed algorithm can be employed for robots with a limited communication range as long as they can configure an ad-hoc network.
To the best of our knowledge, this is the first decentralized MATP algorithm that guarantees the feasibility of the optimization problem, collision avoidance, and deadlock-free in a dense maze-like environment.
We conducted a hardware demonstration to verify the operability of the proposed algorithm, as shown in Fig. \ref{fig: thumbnail}. 
We release the source code in \url{https://github.com/qwerty35/lsc_dr_planner}.

We summarize the main contributions as follows:
\begin{itemize}
\item Decentralized multi-agent trajectory planning algorithm that guarantees to prevent deadlock in a dense maze-like environment.
\item Constraint generation method that ensures the feasibility of the optimization problem and collision avoidance under the limited communication range.
\item Subgoal optimization method that allows the agent to converge to the waypoint without causing a deadlock.
\end{itemize}

\section{RELATED WORK}
\label{sec: related work}
MATP algorithms can be divided into two approaches: centralized and decentralized methods.
The authors of \cite{honig2018trajectory,park2020efficient} present centralized planning algorithms that utilize a grid-based multi-agent path planning (MAPP) algorithm such as ECBS \cite{barer2014suboptimal} to plan an initial trajectory and optimize it. This approach guarantees deadlock-free in a maze-like environment, but it is not scalable to the number of agents.
On the other hand, decentralized methods \cite{van2011reciprocal,luis2020online,zhou2021ego,tordesillas2021mader} show higher scalability than centralized ones, but they often suffer deadlock in a narrow corridor.

For deadlock resolution, many decentralized algorithms adopt the right-hand rule \cite{zhou2017fast,wang2017safety,zhu2019b,abdullhak2021deadlock,toumieh2022decentralized}, which moves the goal point to the right side after the deadlock is detected.
This approach works well in an obstacle-free environment, but there is a risk of another deadlock even after changing the goal point.
Another deadlock resolution method is to replan each robot's trajectory sequentially. In \cite{jager2001decentralized}, a local coordinator asks neighboring agents to plan different trajectories until the deadlock is resolved. 
The authors of \cite{desaraju2012decentralized} introduce a token-based cooperative strategy, that determines which robots to yield the path by bidding.
However, under these methods, there are cases where deadlock cannot be resolved by replanning an alternative trajectory of individual agents. 
The authors of \cite{alonso2018reactive} introduce a centralized high-level coordinator for deadlock resolution. This method is suitable for deadlock resolution in a cluttered environment, but all agents must be connected to the centralized coordinator during the entire mission.

Several works guarantee deadlock-free in obstacle-free or sparse environments.
The authors of \cite{chen2022recursive} introduce a warning band to prevent the agents from clumping together.
In \cite{semnani2020force}, an artificial potential field (APF) is extended to solve the deadlock.
The authors of \cite{grover2020deadlock} conduct deadlock analysis and resolution for 2 to 3 agents.
However, there is a limitation that these methods cannot solve deadlock in a cluttered environment such as a maze.
In \cite{dergachev2021distributed, hou2022enhanced}, the grid-based MAPP is utilized to solve deadlock, similar to the proposed method.
The authors of \cite{dergachev2021distributed} adopt a mode-switching strategy, which converts the planner mode to follow the waypoint from MAPP when the deadlock is detected.
The authors of \cite{hou2022enhanced} utilize the discrete path from MAPP as an initial trajectory.
However, these methods do not provide a theoretical guarantee for deadlock resolution.
Compared to the previous work \cite{park2022online}, the proposed algorithm does not require a fully connected network for collision avoidance, and it guarantees deadlock-free for dense maze-like environments.

\section{PROBLEM STATEMENT}
\label{sec: problem statement}
We suppose that $N$ agents with radius $r$ are deployed in a 2-dimensional space with static obstacles. 
Our goal is to plan a safe and deadlock-free trajectory for the agents under a limited communication range.
The start and goal points of the agent $i$ are $\textbf{s}^{i}$ and $\textbf{g}^{i}_{des}$, respectively.
We denote a set that includes all agents as $\mathcal{I}$ and a set consisting of agent $i$ and the agents that can communicate with the agent $i$ as a \textit{connected group} $\mathcal{N}^{i}$.



\subsection{Assumption}

\subsubsection{Obstacle} The position of the static obstacles is given as prior knowledge. 
\subsubsection{Grid-based planner} 
All agents share the same grid space $G=(V,E)$, where the grid size $d$ is larger than $2\sqrt{2}r$. If the agent is on the grid, then the agent does not collide with static obstacles.
\subsubsection{Mission} The start and desired goal points of all agents are located at the vertex of the grid space, and inter-agent collision do not occur at the start point. All agents start the mission at the same time. 
\subsubsection{Communication} The agents $i$ and $j$ can share the information without a communication loss or delay if the agents satisfy the following: 
\begin{equation}
    \|\textbf{p}^{i}(t) - \textbf{p}^{j}(t)\|_{\infty} \leq r_{c}
\label{eq: communication range}
\end{equation}
where $\textbf{p}^{i}(t)$ is the position of the agent $i$, $\|\cdot\|_{\infty}$ is the L-infinity norm, and $r_{c} > 2d$ is the communication range. All agents can configure an ad-hoc network to relay information between the agents within the communication range.

\subsection{Agent Description}
\subsubsection{Trajectory representation}
We represent the agent's trajectory to the $M$-segment piecewise Bernstein polynomial \cite{farouki2012bernstein}, thanks to the differential flatness of quadrotor dynamics \cite{mellinger2012mixed}. More precisely, the $m^{th}$ segment of the trajectory of the agent $i$ is formulated as follows: 
\begin{equation}
    \textbf{p}^{i}_{k}(t) = \sum_{l=0}^{n}\textbf{c}^{i}_{k,m,l} b_{l,n}(\frac{t-T_{k+m-1}}{\Delta t}), \forall t \in [T_{k+m-1},T_{k+m}]
\label{eq: trajectory representation}
\end{equation}
where $k$ is the current replanning step, $\textbf{p}^{i}_{k}(t)$ is the trajectory of the agent $i$, $\textbf{c}^{i}_{k,m,l} \in \mathbb{R}^{2}$ is the control point, $n>4$ is the degree of the polynomial, $b_{l,n}(t)$ is Bernstein basis polynomial, $T_{0}$ is the mission start time, $T_{k} = T_{0} + k \Delta t$, and $\Delta t$ is the duration of the trajectory segment. 

\subsubsection{Collision avoidance}
We define that the agent $i$ is safe from a collision if the following conditions hold: 
\begin{equation}
    \|\textbf{p}^{i}_{k}(t) - \textbf{p}^{j}_{k}(t)\| \geq 2r, \forall t, j \in \mathcal{I} \backslash \{i\}
\label{eq: definition of inter-agent avoidance}
\end{equation}
\begin{equation}
    (\textbf{p}^{i}_{k}(t) \oplus \mathcal{C}^{i,o}) \cap \mathcal{O} = \emptyset, \forall t
\label{eq: definition of obstacle avoidance}
\end{equation}
\begin{equation}
  \mathcal{C}^{i,o}=\{\textbf{x} \in \mathbb{R}^{2} \mid \|\textbf{x}\| < r\}
\label{eq: obstacle collision model}
\end{equation}
where $\oplus$ is the Minkowski sum, $\mathcal{C}^{i,o}$ is the obstacle collision model, $\mathcal{O}$ is the space occupied by the obstacles, and $\|\cdot\|$ is the Euclidean norm.

\subsubsection{Dynamical limit}
We model the dynamical limit of the agent as follows:
\begin{equation}
    \|\textbf{v}^{i}(t)\|_{\infty} \leq v_{max}, \forall t
\label{eq: maximum velocity}
\end{equation}
\begin{equation}
    \|\textbf{a}^{i}(t)\|_{\infty} \leq a_{max}, \forall t 
\label{eq: maximum acceleration}
\end{equation}
where $\textbf{v}^{i}(t)$ and $\textbf{a}^{i}(t)$ are the velocity and acceleration of the agent $i$, respectively, and $v_{max}$ and $a_{max}$ are the agent's maximum velocity and acceleration, respectively. 






\section{METHOD}

As described in Alg. \ref{alg: trajectory planning algorithm}, the proposed algorithm consists of the communication phase (lines 3-4) and trajectory generation phase (lines 5-13).
During the communication phase, each agent configures an ad-hoc network between the agents within the communication range. 
After network configuration, we conduct a grid-based multi-agent path planning (MAPP) algorithm to determine the waypoint of the agent (line 3, Sec. \ref{subsec: decentralized MAPP}). Then, the agent shares the previously planned trajectory and subgoal with the connected group (line 4).
In the trajectory generation phase, we generate initial trajectories using the previously planned trajectories (line 6, Sec. \ref{subsec: initial trajectory planning}).
The initial trajectories are utilized to construct feasible collision constraints (lines 7-9, Sec. \ref{subsec: collision constraints construction}).
Next, we search for the subgoal that does not cause deadlock (line 10, Sec. \ref{subsec: subgoal optimization}).
Finally, we conduct trajectory optimization and execute it (lines 11-12, Sec. \ref{subsec: trajectory optimization}).
We repeat the above process until all agents reach the desired goal. 

\begin{algorithm}
\SetAlgoLined
\KwIn{
Start point $\textbf{s}^{i}$, desired goal point $\textbf{g}^{i}_{des}$, and obstacle space $\mathcal{O}$}
\KwOut{Trajectory of the agent $i$ $\textbf{p}^{i}_{k}(t)$}
  $k \gets 0$\;
  \While{not all agents at desired goal}{
    \tcp{Communication phase}
    $\textbf{w}^{j \in \mathcal{N}^{i}}_{k} \gets$ decentralizedMAPP($\textbf{g}^{i}_{k-1}, \textbf{w}^{i}_{k-1},\textbf{g}^{i}_{des}$)\;
    $\textbf{p}^{j \in \mathcal{N}^{i}}_{k-1}(t), \textbf{g}^{j \in \mathcal{N}^{i}}_{k-1} \gets$ communicate($\textbf{p}^{i}_{k-1}(t), \textbf{g}^{i}_{k-1}$)\;
    \tcp{Trajectory generation phase}
    \For{$\forall j \in \mathcal{N}^{i}$}{
      $\hat{\textbf{p}}^{j}_{k}(t) \gets$ planInitialTraj($\textbf{p}^{j}_{k-1}(t)$)\;
      $\mathcal{L}^{i,j}_{k,m,l} \gets$ buildLSC($\hat{\textbf{p}}^{i}_{k}(t), \hat{\textbf{p}}^{j}_{k}(t)$)\;
    }
    $\mathcal{S}^{i}_{k,m} \gets$ buildSFC($\hat{\textbf{p}}^{i}_{k}(t), \mathcal{O}$)\;
    $\textbf{g}^{i}_{k} \gets$ subgoalOpt($\textbf{g}^{i}_{k-1}, \textbf{w}^{i}_{k}, \mathcal{S}^{i}_{k,m}, \mathcal{L}^{i,j}_{k,m,l}$)\;
    $\textbf{p}^{i}_{k}(t) \gets$ trajOpt($\mathcal{S}^{i}_{k,m}, \mathcal{L}^{i,j}_{k,m,l}, \textbf{g}^{i}_{k})$\;
    executeTrajectory($\textbf{p}^{i}_{k}(t)$)\;
    $k \gets k+1$\; 
  }
\caption{Trajectory planning for the agent $i$}
\label{alg: trajectory planning algorithm}
\end{algorithm}

\subsection{Decentralized Multi-agent Path Planning}
\label{subsec: decentralized MAPP}

\begin{algorithm}
\SetAlgoLined
\KwIn{Prev. subgoals $\textbf{g}^{j\in\mathcal{N}^{i}}_{k-1}$, prev. waypoints $\textbf{w}^{j\in\mathcal{N}^{i}}_{k-1}$, desired goals $\textbf{g}^{j\in\mathcal{N}^{i}}_{des}$}
\KwOut{Current waypoints $\textbf{w}^{j \in \mathcal{N}^{i}}_{k}$}
  \tcp{Update waypoints}
  $\boldsymbol{\pi}^{j\in\mathcal{N}^{i}} \gets$ runMAPP($\textbf{w}^{j\in\mathcal{N}^{i}}_{k-1}, \textbf{g}^{j\in\mathcal{N}^{i}}_{des}$)\;
  $\mathcal{Q} \gets \emptyset$\;
  \For{$\forall j \in \mathcal{N}^{i}$}{
    
    \uIf{$k > 0$ and the agent $j$ satisfies (\ref{eq: subgoal update condition1}), (\ref{eq: subgoal update condition2}))}{
      $\mathcal{Q} \gets \mathcal{Q} \cup \{j\}$\;
      $\textbf{w}^{j}_{k} \gets \boldsymbol{\pi}^{j}[1]$\;
    }
    \Else{
      $\textbf{w}^{j}_{k} \gets \textbf{w}^{j}_{k-1}$\;
    }
  }
  \tcp{Conflict resolution}
  \For{$\forall j \in \mathcal{Q}$}{
    \If{$\textbf{w}^{j}_{k} = \textbf{w}^{q}_{k}, \exists q \in \mathcal{N}^{i}\backslash\{j\}$}{
      $\textbf{w}^{j}_{k} \gets \textbf{w}^{j}_{k-1}$\;
    }
  }
  \KwRet{$\textbf{w}^{j\in\mathcal{N}^{i}}_{k}$}
\caption{decentralizedMAPP}
\label{alg: decentralized MAPP}
\end{algorithm}

We introduce a decentralized MAPP to plan the waypoint, which provides guidance on deadlock resolution. 
Alg. \ref{alg: decentralized MAPP} describes the proposed waypoint update method.
For every replanning step, each agent configures the ad-hoc network between agents within the communication range, and one agent among the connected group is selected as a local coordinator. The local coordinator collects the subgoals, waypoints, and desired goals of the agents in the connected group. Then, the coordinator plans collision-free discrete paths using the MAPP algorithm on the grid space $G$, where the start points of MAPP are the previous waypoints $\textbf{w}^{i \in \mathcal{N}^{i}}_{k-1}$, and the goal points are the desired goals (line 1). 
If it is the first step of the planning, we set the start point as $\textbf{s}^{i}$ instead. In this work, we adopt Priority Inheritance with Backtracking (PIBT) \cite{okumura2022priority} for MAPP algortihm because it is a scalable algorithm that guarantees \textit{reachability}, which ensures that all agents can reach the desired goal within a finite time.
Next, the coordinator updates the agent's waypoint $\textbf{w}^{i}_{k}$ to the second waypoint of the discrete path (the point one step after the start point) if the following two conditions are satisfied (lines 3-6). 
First, the subgoal and waypoint at the previous step must be equal (\ref{eq: subgoal update condition1}). 
Second, the distance between the updated waypoint and the endpoints of the previous trajectory's segments must be shorter than $r_{c}/2$ (\ref{eq: subgoal update condition2}):
\begin{equation}
  \textbf{g}^{i}_{k-1} = \textbf{w}^{i}_{k-1}
\label{eq: subgoal update condition1}
\end{equation}
\begin{equation}
  \|\textbf{w}^{i}_{k} - \textbf{p}^{i}_{k-1}(T_{k+m-2})\|_{\infty} < \frac{r_{c}}{2}, \forall m=1,\cdots,M
\label{eq: subgoal update condition2}
\end{equation}
where $\textbf{g}^{i}_{k}$ and $\textbf{w}^{i}_{k}$ are the subgoal and waypoint at the replanning step $k$, respectively. 
Otherwise, we reuse the previous waypoint as the current waypoint (lines 7-9).
Lastly, we check whether the waypoints are duplicated in the connected group. If there are the same ones, we restore one of them to the previous waypoint. We repeat this process until there is no duplicated waypoint (line 11-15). 
Lemma \ref{lemma: no duplicated waypoints} shows that the proposed waypoint update rule prevents the duplicated waypoints.
\begin{lemma}
For any pair of the agents $i \in \mathcal{I}$ and $j \in \mathcal{I} \backslash \{i\}$, $\textbf{w}^{i}_{k} \neq \textbf{w}^{j}_{k}$ holds for every replanning step $k > 0$.
\label{lemma: no duplicated waypoints}
\end{lemma}

\begin{proof}
If $j \in \mathcal{N}^{i}$, then the waypoints of the agent $i$ and $j$ cannot be duplicated because we eliminate the duplicated waypoints at the lines 11-15 in Alg. \ref{alg: decentralized MAPP}.
Assume that $j \notin \mathcal{N}^{i}$ and the agents $i$ and $j$ have the same waypoints $\textbf{w}^{i}_{k} = \textbf{w}^{j}_{k}$.
We have $\textbf{p}^{i}_{k-1}(T_{k+1}) = \textbf{p}^{i}_{k}(T_{k})$ by the initial condition of the trajectory. Also, we obtain the following due to (\ref{eq: subgoal update condition2}) and (\ref{eq: communication range2}):  
\begin{equation}
    \|\textbf{w}^{i}_{k} - \textbf{p}^{i}_{k}(T_{k})\|_{\infty} < \frac{r_{c}}{2}
\label{eq: lemma1_1}
\end{equation}
\begin{equation}
    \|\textbf{w}^{j}_{k} - \textbf{p}^{j}_{k}(T_{k})\|_{\infty} < \frac{r_{c}}{2}
\label{eq: lemma1_2}
\end{equation}
Therefore, the distance between two agents are smaller than the communication range by triangle inequality:
\begin{equation}
    \|\textbf{p}^{i}_{k}(T_{k}) - \textbf{p}^{j}_{k}(T_{k})\|_{\infty} < r_{c}
\label{eq: lemma1_3}
\end{equation}
However, it is inconsistent with the assumption that $j \notin \mathcal{N}^{i}$. Thus, there is no duplicated waypoints.
\end{proof}

\subsection{Initial Trajectory Planning}
\label{subsec: initial trajectory planning}
As in our previous work \cite{park2022online}, we utilize an initial trajectory to construct feasible collision constraints. We plan the initial trajectory using the previously planned trajectories, or the current position if it is the first step of the planning:
\begin{equation}
\begin{alignedat}{2}
    \hat{\textbf{p}}^{i}_{k}(t) = 
    \begin{cases} 
    \ \textbf{s}^{i} & k = 0, t \in [T_{0}, T_{M}] \\  
    \ \textbf{p}^{i}_{k-1}(t)   & k > 0, t \in [T_{k}, T_{k+M-1}] \\    
    \ \textbf{p}^{i}_{k-1}(T_{k+M-1})   & k > 0, t \in [T_{k+M-1}, T_{k+M}] \\
    \end{cases}
\end{alignedat}
\label{eq: initial trajectory}
\end{equation}
where $\hat{\textbf{p}}^{i}_{k}(t)$ is the initial trajectory at the replanning step $k$. The control point of the initial trajectory is represented as follows:
\begin{equation}
\begin{alignedat}{2}
    \hat{\textbf{c}}^{i}_{k,m,l} = 
    \begin{cases} 
    \ \textbf{s}^{i}   & k = 0 \\  
    \ \textbf{c}^{i}_{k-1,m+1,l}   & k > 0, m < M \\    
    \ \textbf{c}^{i}_{k-1,M,n}   & k > 0, m = M \\
    \end{cases}
\end{alignedat}
\label{eq: control point of initial trajectory}
\end{equation}
where $\hat{\textbf{c}}^{i}_{k,m,l}$ is the control point of the initial trajectory.

\subsection{Collision Constraints Construction}
\label{subsec: collision constraints construction}
In our previous work \cite{park2022online}, we utilized a safe flight corridor (SFC) and linear safe corridor (LSC) for collision avoidance. However, these constraints may cause deadlock if the agent is blocked by the constraints before reaching the waypoint. For this reason, we modify the collision constraints so that the agent can proceed to the waypoint.

\subsubsection{Obstacle avoidance}
For obstacle avoidance, we construct the SFC as follows:
\begin{equation}
\begin{alignedat}{2}
    \mathcal{S}^{i}_{k,m} = 
    \begin{cases} 
    \ \mathcal{S}(\{\textbf{s}^{i}, \textbf{w}^{i}_{k}\})   & k = 0\\  
    \ \mathcal{S}^{i}_{k-1,m+1}   & k > 0, m < M \\
    \ \mathcal{S}(\{\hat{\textbf{c}}^{i}_{k,M,n},\textbf{g}^{i}_{k-1},\textbf{w}^{i}_{k}\}))   & k > 0, m = M, (\ref{eq: safe flight corridor condition}) \\
    \ \mathcal{S}(\{\hat{\textbf{c}}^{i}_{k,M,n},\textbf{g}^{i}_{k-1}\})   & \text{else} \\
    \end{cases}
\end{alignedat}
\label{eq: safe flight corridor construction}
\end{equation}
\begin{equation}
    (\text{Conv}(\{\hat{\textbf{c}}^{i}_{k,M,n}, \textbf{g}^{i}_{k-1},\textbf{w}^{i}_{k}\}) \oplus \mathcal{C}^{i,o}) \cap \mathcal{O} = \emptyset
\label{eq: safe flight corridor condition}
\end{equation}
where $\mathcal{S}^{i}_{k,m}$ is the SFC for $m^{th}$ trajectory segment, $\mathcal{S}(\mathcal{P})$ is a convex set that includes the point set $\mathcal{P}$ and satisfies $(\mathcal{S}(\mathcal{P}) \oplus \mathcal{C}^{i,o}) \cap \mathcal{O} = \emptyset$, and $\text{Conv}(\cdot)$ is the convex hull operator that returns a convex hull of input set. We generate the SFC using the axis-search method \cite{park2020efficient}.


\subsubsection{Inter-agent collision avoidance}
If it is the first step of the planning or $m<M$, we construct the LSC using the same approach in \cite{park2022online}:
\begin{subequations}
\begin{gather}
        \mathcal{L}^{i,j}_{k,m,l} = \{\textbf{x} \in \mathbb{R}^{2} \mid (\textbf{x} - \hat{\textbf{c}}^{j}_{k,m,l}) \cdot \textbf{n}^{i,j}_{m} - d^{i,j}_{m,l} \geq 0\}  \\
    d^{i,j}_{m,l} = r + \frac{1}{2}(\hat{\textbf{c}}^{i}_{k,m,l} - \hat{\textbf{c}}^{j}_{k,m,l}) \cdot \textbf{n}^{i,j}_{m} 
\end{gather}
\label{eq: linear safe corridor}%
\end{subequations}
where $\mathcal{L}^{i,j}_{k,m,l}$ is the LSC between the agent $i$ and $j$, $\textbf{n}^{i,j}_{m}$ is the normal vector that satisfies $\textbf{n}^{i,j}_{m} = -\textbf{n}^{j,i}_{m}$, $d^{i,j}_{m,l}$ is the safety margin. The detailed LSC construction can be found in \cite{park2022online}.
If it is not the first step of the planning and $m = M$, then we generate the LSC as follows:
\begin{subequations}
\begin{gather}
      \mathcal{L}^{i,j}_{k,M,l} = \{\textbf{x} \in \mathbb{R}^{2} \mid (\textbf{x} - \textbf{p}^{j}_{cls,i}) \cdot \textbf{n}^{i,j}_{M} - d^{i,j}_{M,l} \geq 0\} \\
      \textbf{n}^{i,j}_{M} = \frac{\textbf{p}^{i}_{cls,j}-\textbf{p}^{j}_{cls,i}}{\|\textbf{p}^{i}_{cls,j}-\textbf{p}^{j}_{cls,i}\|} \\
      d^{i,j}_{M,l} = r + \frac{1}{2}\|\textbf{p}^{i}_{cls,j}-\textbf{p}^{j}_{cls,i}\|
\end{gather}
\label{eq: linear safe corridor last segment}%
\end{subequations}
where $\textbf{p}^{i}_{cls,j} \in \langle \hat{\textbf{c}}^{i}_{k,M,n}, \textbf{g}^{i}_{k-1} \rangle$ and $\textbf{p}^{j}_{cls,i} \in \langle \hat{\textbf{c}}^{j}_{k,M,n}, \textbf{g}^{j}_{k-1} \rangle$ are the closest points between $\langle \hat{\textbf{c}}^{i}_{k,M,n}, \textbf{g}^{i}_{k-1} \rangle$ and $\langle \hat{\textbf{c}}^{j}_{k,M,n}, \textbf{g}^{j}_{k-1} \rangle$, respectively, and $\langle \textbf{a},\textbf{b} \rangle$ is the line segment between two points $\textbf{a}$ and $\textbf{b}$. 

\begin{figure}[t]
\centering
\includegraphics[width = 0.7\linewidth]{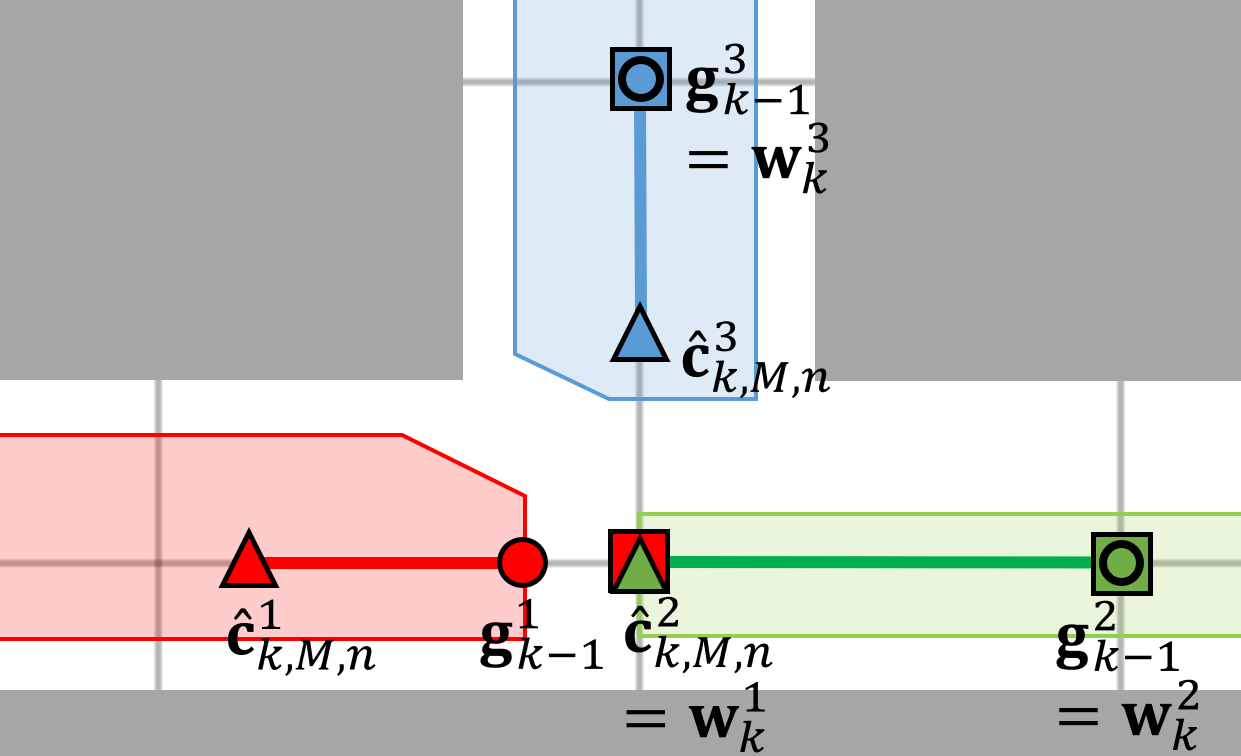}
\caption{
Collision constraints for the last trajectory segment. The squares are the waypoints, the triangles are the final points of the initial trajectories, and the circles are the previously planned subgoals. The gray box is the static obstacle, and the color-shaded region is the feasible region that satisfies the collision constraints. We generate the collision constraint for the last segment to include the line segment between the final point and the subgoal, which is depicted as the thick line.
}
\label{fig: collision constraints}
\end{figure}

Fig. \ref{fig: collision constraints} shows the collision constraints for the last trajectory segment. We can observe that the feasible region of the agent always contains $\langle \textbf{g}^{i}_{k-1},\hat{\textbf{c}}^{i}_{k,M,n} \rangle$. Thus, each agent can secure the free space to proceed to the subgoal $\textbf{g}^{i}_{k-1}$, which will converge to the waypoint $\textbf{w}^{i}_{k}$.

\subsection{Subgoal Optimization}
\label{subsec: subgoal optimization}

Suppose that the waypoint from MAPP does not satisfy the collision constraints.
If we directly set this waypoint as the target point, this may lead to deadlock since the agent cannot reach the waypoint by the constraints.
Therefore, we designate the point that is closest to the waypoint and satisfies the collision constraints as the subgoal. More precisely, we determine the subgoal by solving the following linear programming (LP) problem:

\begin{equation}
\begin{aligned}
& \underset{\textbf{g}^{i}_{k}}{\text{minimize}}     & & \|\textbf{g}^{i}_{k} - \textbf{w}^{i}_{k}\| \\
& \text{subject to}   & & \textbf{g}^{i}_{k} \in \langle \textbf{s}^{i},\textbf{w}^{i}_{k}\rangle & & \text{if } k = 0 \\
&                     & & \textbf{g}^{i}_{k} \in \langle \textbf{g}^{i}_{k-1},\textbf{w}^{i}_{k}\rangle & & \text{if } k > 0  \\
&                     & & \textbf{g}^{i}_{k} \in \mathcal{S}^{i}_{k,M}  \\
&                     & & \textbf{g}^{i}_{k} \in \mathcal{L}^{i,j}_{k,M,n}  & & \forall j \in \mathcal{N}^{i} \\
\end{aligned}
\label{eq: subgoal optimization}
\end{equation}
where $\textbf{g}^{i}_{k}$ is the subgoal at the replanning step $k$. 
We will prove that the subgoal in (\ref{eq: subgoal optimization}) does not cause deadlock in section \ref{sec: theoretical guarantee}.
Lemma \ref{lemma: one agent per edge} shows the properties of the subgoal.
\begin{lemma}
 For the agents $i \in \mathcal{I}$, $j \in \mathcal{I} \backslash \{i\}$,
 (i) there exists a grid edge $e \in E$ such that $\langle \textbf{g}^{i}_{k}, \textbf{w}^{i}_{k} \rangle \subset e$, (ii) $\textbf{g}^{i}_{k} \neq \textbf{g}^{j}_{k}$, (iii) if there exists an edge $e \in E$ such that $\textbf{g}^{i}_{k} \in e$ and $\textbf{g}^{j}_{k} \in e$, then $\textbf{g}^{i}_{k}$ or $\textbf{g}^{j}_{k}$ is on the vertex of the grid $G=(V,E)$.
\label{lemma: one agent per edge}
\end{lemma}

\begin{proof}
(i) If $k=0$, there exists a grid edge $e \in E$ such that $\langle \textbf{s}^{i}, \textbf{w}^{i}_{0} \rangle \subset e$ because $\textbf{s}^{i}$ are located at the vertex by the assumption and $\textbf{w}^{i}_{0}$ is the waypoint of discrete path from MAPP. Therefore, we obtain $\langle \textbf{g}^{i}_{0}, \textbf{w}^{i}_{0} \rangle \subset e$
since $\textbf{g}^{i}_{0} \in \langle \textbf{s}^{i}, \textbf{w}^{i}_{0} \rangle$ by (\ref{eq: subgoal optimization}).

Assume that there exists a grid edge $e_{k-1} \in E$ such that $\langle \textbf{g}^{i}_{k-1}, \textbf{w}^{i}_{k-1} \rangle \subset e_{k-1}$.
If $\textbf{g}^{i}_{k-1} \neq \textbf{w}^{i}_{k-1}$, then we have $\textbf{w}^{i}_{k} = \textbf{w}^{i}_{k-1}$ by the waypoint update rule (\ref{eq: subgoal update condition1}).
Hence we obtain $\langle \textbf{g}^{i}_{k}, \textbf{w}^{i}_{k} \rangle \subset \langle \textbf{g}^{i}_{k-1}, \textbf{w}^{i}_{k} \rangle \subset e_{k-1}$ since $\textbf{g}^{i}_{k} \in \langle \textbf{g}^{i}_{k-1}, \textbf{w}^{i}_{k} \rangle$ by (\ref{eq: subgoal optimization}). 
If $\textbf{g}^{i}_{k-1} = \textbf{w}^{i}_{k-1}$, then there exists a grid edge $e$ such that $\langle \textbf{g}^{i}_{k}, \textbf{w}^{i}_{k} \rangle = \langle \textbf{w}^{i}_{k-1}, \textbf{w}^{i}_{k} \rangle \subset e$ because $\textbf{w}^{i}_{k-1}, \textbf{w}^{i}_{k}$ are the consecutive waypoints of the discrete path.

Thus, there exists a grid edge $e \in E$ such that $\langle \textbf{g}^{i}_{k}, \textbf{w}^{i}_{k} \rangle \subset e$ for every replanning step $k$ by mathematical induction.

(ii) We have $\|\textbf{g}^{i}_{k} - \textbf{g}^{j}_{k}\| \geq 2r$ for every replanning step $k$ because $\textbf{g}^{i}_{k} \in \mathcal{L}^{i,j}_{k,M,n}$ and $\textbf{g}^{j}_{k} \in \mathcal{L}^{j,i}_{k,M,n}$.
It implies $\textbf{g}^{i}_{k} \neq \textbf{g}^{j}_{k}$.

(iii) Let us define $(\textbf{a},\textbf{b})$ as follows:
\begin{equation}
\begin{alignedat}{2}
    (\textbf{a},\textbf{b}) = 
    \begin{cases} 
    \ \emptyset   & \text{if } \textbf{a} = \textbf{b} \\  
    \ \{\textbf{a} + \alpha (\textbf{b} - \textbf{a}) \mid 0 < \alpha < 1  \}   & \text{if } \textbf{a} \neq \textbf{b} \\    
    \end{cases}
\end{alignedat}
\label{eq: line segment except vertex}
\end{equation}
For the proof, we first show that the Alg. \ref{alg: trajectory planning algorithm} satisfies the following using the mathematical induction:
\begin{equation}
(\textbf{g}^{i}_{k-1}, \textbf{w}^{i}_{k}) \cap (\textbf{g}^{j}_{k-1}, \textbf{w}^{j}_{k}) = \emptyset, \forall k > 0
\label{eq: no duplicated edge1}
\end{equation}
\begin{equation}
(\textbf{g}^{i}_{k}, \textbf{w}^{i}_{k}) \cap (\textbf{g}^{j}_{k}, \textbf{w}^{j}_{k}) = \emptyset, \forall k
\label{eq: no duplicated edge2}
\end{equation}
If $k = 0$, we have $(\textbf{s}^{i}, \textbf{w}^{i}_{0}) \cap (\textbf{s}^{j}, \textbf{w}^{j}_{0}) = \emptyset$ because we plan the waypoints using MAPP that ensures collision avoidance.
We thus obtain $(\textbf{g}^{i}_{0}, \textbf{w}^{i}_{0}) \cap ( \textbf{g}^{j}_{0}, \textbf{w}^{j}_{0}) = \emptyset$ since $(\textbf{g}^{i}_{0}, \textbf{w}^{i}_{0}) \subset (\textbf{s}^{i}, \textbf{w}^{i}_{0})$ by (\ref{eq: subgoal optimization}).

Assume that $(\textbf{g}^{i}_{k-1}, \textbf{w}^{i}_{k-1}) \cap (\textbf{g}^{j}_{k-1}, \textbf{w}^{j}_{k-1}) = \emptyset$. 
\newline
(Case 1) If $\textbf{g}^{i}_{k-1} \neq \textbf{w}^{i}_{k-1}$ and $\textbf{g}^{j}_{k-1} \neq \textbf{w}^{j}_{k-1}$, we have $(\textbf{g}^{i}_{k-1}, \textbf{w}^{i}_{k}) \cap (\textbf{g}^{j}_{k-1}, \textbf{w}^{j}_{k}) = \emptyset$ since $\textbf{w}^{i}_{k-1} = \textbf{w}^{i}_{k}$ and $\textbf{w}^{j}_{k-1} = \textbf{w}^{j}_{k}$ by the waypoint update rule (\ref{eq: subgoal update condition1}). Therefore, we obtain $(\textbf{g}^{i}_{k}, \textbf{w}^{i}_{k}) \cap (\textbf{g}^{j}_{k}, \textbf{w}^{j}_{k}) = \emptyset$ by (\ref{eq: subgoal optimization}). 
\newline
(Case 2) If $\textbf{g}^{i}_{k-1} = \textbf{w}^{i}_{k-1}$ and $\textbf{g}^{j}_{k-1} = \textbf{w}^{j}_{k-1}$, then we have $(\textbf{g}^{i}_{k-1}, \textbf{w}^{i}_{k}) \cap (\textbf{g}^{j}_{k-1}, \textbf{w}^{j}_{k}) = (\textbf{w}^{i}_{k-1}, \textbf{w}^{i}_{k}) \cap (\textbf{w}^{j}_{k-1}, \textbf{w}^{j}_{k}) = \emptyset$ since Alg. \ref{alg: decentralized MAPP} updates the waypoints without any conflict. 
Therefore, we obtain $(\textbf{g}^{i}_{k}, \textbf{w}^{i}_{k}) \cap (\textbf{g}^{j}_{k}, \textbf{w}^{j}_{k}) = \emptyset$ because $(\textbf{g}^{i}_{k}, \textbf{w}^{i}_{k}) \subset (\textbf{g}^{i}_{k-1}, \textbf{w}^{i}_{k})$ by (\ref{eq: subgoal optimization}). 
\newline
(Case 3) Assume that only one of the subgoals is equal to the waypoint. In other words, $\textbf{g}^{i}_{k-1} \neq \textbf{w}^{i}_{k-1}$ and $\textbf{g}^{j}_{k-1} = \textbf{w}^{j}_{k-1}$ without loss of generality. Then, we have $(\textbf{g}^{i}_{k-1}, \textbf{w}^{i}_{k}) \cap (\textbf{g}^{j}_{k-1}, \textbf{w}^{j}_{k}) = (\textbf{g}^{i}_{k-1}, \textbf{w}^{i}_{k-1}) \cap (\textbf{w}^{j}_{k-1}, \textbf{w}^{j}_{k})$ by the waypoint update rule (\ref{eq: subgoal update condition1}). 
Here, we can find a grid edge $e_{k-1} \in E$ such that $(\textbf{g}^{i}_{k-1}, \textbf{w}^{i}_{k-1}) \subset e_{k-1}$ by (i) of Lemma \ref{lemma: one agent per edge}. Also, it satisfies $(\textbf{w}^{j}_{k-1}, \textbf{w}^{j}_{k}) \cap e_{k-1} = \emptyset$ because $\textbf{w}^{j}_{k-1} \neq \textbf{w}^{i}_{k-1}$ and $\textbf{w}^{j}_{k} \neq \textbf{w}^{i}_{k} = \textbf{w}^{i}_{k-1}$ by Lemma \ref{lemma: no duplicated waypoints}.
Therefore, we obtain $(\textbf{g}^{i}_{k-1}, \textbf{w}^{i}_{k}) \cap (\textbf{g}^{j}_{k-1}, \textbf{w}^{j}_{k}) = \emptyset$, which implies that $(\textbf{g}^{i}_{k}, \textbf{w}^{i}_{k}) \cap (\textbf{g}^{j}_{k}, \textbf{w}^{j}_{k}) = \emptyset$ by (\ref{eq: subgoal optimization}). 
Thus, $(\textbf{g}^{i}_{k}, \textbf{w}^{i}_{k}) \cap (\textbf{g}^{j}_{k}, \textbf{w}^{j}_{k}) = \emptyset$ for every replanning step by mathematical induction.

Next, we will show that $\textbf{g}^{i}_{k} \in V$ or $\textbf{g}^{j}_{k} \in V$ when $\textbf{g}^{i}_{k}$ and $\textbf{g}^{j}_{k}$ are on the same edge $e \in E$ using the mathematical induction. 
Assume that there exists agents $i$ and $j$ such that $\textbf{g}^{i}_{0} \notin V$, $\textbf{g}^{j}_{0} \notin V$ when $\textbf{g}^{i}_{0} \in e$, $\textbf{g}^{j}_{0} \in e$. Since $\textbf{s}^{i}, \textbf{s}^{j}, \textbf{w}^{i}_{0}, \textbf{w}^{j}_{0}$ are on the vertex, the agents must satisfy the following to hold $\textbf{g}^{i}_{0} \notin V$, $\textbf{g}^{j}_{0} \notin V$:
\begin{subequations}
\begin{align}
    & (\textbf{s}^{i},\textbf{w}^{i}_{0}) \cap e \neq \emptyset \\
     & (\textbf{s}^{j},\textbf{w}^{j}_{0}) \cap e \neq \emptyset  
\end{align}
\label{eq: lemma2_1}%
\end{subequations}
However, it means that there exists a collision between discrete paths from the MAPP. Thus, $\textbf{g}^{i}_{0} \in V$ or $\textbf{g}^{j}_{0} \in V$.

Assume that (iii) of Lemma \ref{lemma: one agent per edge} is true at the replanning step $k-1$. 
Suppose that $\textbf{g}^{i}_{k} \notin V$, $\textbf{g}^{j}_{k} \notin V$ when $\textbf{g}^{i}_{k} \in e$, $\textbf{g}^{j}_{k} \in e$, the agents $i$ and $j$ must satisfy the following to hold  by (\ref{eq: subgoal optimization}):
\begin{subequations}
\begin{align}
    & \langle \textbf{g}^{i}_{k-1},\textbf{w}^{i}_{k} \rangle \cap e \neq \emptyset  \\
    & \langle \textbf{g}^{j}_{k-1},\textbf{w}^{j}_{k} \rangle \cap e \neq \emptyset 
\end{align}
\label{eq: lemma2_2}%
\end{subequations}
However, it is impossible to satisfy the above conditions because the necessary conditions of (\ref{eq: lemma2_2}) become $\textbf{g}^{i}_{k-1} \notin V$ and $\textbf{g}^{j}_{k-1} \notin V$ due to (\ref{eq: no duplicated edge1}), which are inconsistent with the assumption that Lemma \ref{lemma: one agent per edge} is true at the replanning step $k-1$.
Therefore, we obtain $\textbf{g}^{i}_{k} \notin V$ or $\textbf{g}^{j}_{k} \notin V$.
In conclusion, if there exists an edge $e \in E$ such that $\textbf{g}^{i}_{k} \in e$ or $\textbf{g}^{j}_{k} \in e$, then $\textbf{g}^{i}_{k} \in V$ or $\textbf{g}^{j}_{k} \in V$ by the mathematical induction.
\end{proof}

\subsection{Trajectory Optimization}
\label{subsec: trajectory optimization}

\subsubsection{Objective function}
We formulate the objective function to minimize both the distance to the current subgoal and the jerk of the trajectory as follows:
\begin{equation}
    J^{i}_{err} = w_{err} \|\textbf{p}^{i}_{k}(T_{M})-\textbf{g}^{i}_{k}\|^{2}
\label{eq: objective function error to goal}
\end{equation}
\begin{equation}
\begin{aligned}
    J^{i}_{der} = w_{der}\int_{T_0}^{T_M}\left\|\frac{d^{3}}{dt^{3}}\textbf{p}^{i}_{k}(t)\right\|^{2}dt
\label{eq: objective function derivative of trajectory}
\end{aligned}
\end{equation}
where $w_{err}, w_{der} > 0$ are the weight coefficients.

\subsubsection{Communication range}
If we do not consider the communication range when generating the trajectory, the agent may collide with an agent outside the range. 
Also, if the distance between the agent and its waypoint is longer than half the communication range, an agent outside the range can assign the same waypoint. Hence we add the following constraints to prevent the collision and duplicated waypoints between agents outside the range:
\begin{equation}
    \|\textbf{c}^{i}_{k,m+h,l} - \textbf{c}^{i}_{k,m,0}\|_{\infty} \leq \frac{r_{c}}{2} - r, \forall h \geq 0, m, l, 
\label{eq: communication range1}
\end{equation}
\begin{equation}
    \|\textbf{c}^{i}_{k,m,n} - \textbf{w}^{i}_{k}\|_{\infty} \leq \frac{r_{c}}{2}, \forall m
\label{eq: communication range2}
\end{equation}

\subsubsection{Other constraints}
The trajectory must satisfy the initial condition to match the agent's current state, and we impose continuity constraints to make the trajectory continuous up to the acceleration. 
We add the final stop condition for the feasibility of the optimization problem (i.e., $\textbf{c}^{i}_{k,M,n} = \textbf{c}^{i}_{k,M,n-1} = \textbf{c}^{i}_{k,M,n-2}$).
The dynamical limit (\ref{eq: maximum velocity}), (\ref{eq: maximum acceleration}) can be represented to affine inequality using the convex hull property of the Bernstein polynomial. We can reformulate the above constraints as the following affine constraints:
\begin{equation}
    A_{eq}\textbf{c}^{i}_{k} = \textbf{b}_{eq}
\label{eq: equality constraints}
\end{equation}
\begin{equation}
    A_{dyn}\textbf{c}^{i}_{k} \preceq \textbf{b}_{dyn}
\label{eq: dynamic feasible constraints}
\end{equation}
where $\textbf{c}^{i}_{k}$ is the vector that concatenates the control points of $\textbf{p}^{i}_{k}(t)$.

\subsubsection{Optimization problem}
We conduct the trajectory optimization by solving the following quadratic programming (QP) problem:
\begin{equation}
\begin{aligned}
& \underset{\textbf{c}^{i}_{k}}{\text{minimize}}     & & J^{i}_{err} + J^{i}_{der} \\
& \text{subject to}   & & \textbf{c}^{i}_{k,m,l} \in \mathcal{S}^{i}_{k,m} & & \forall m, l \\
&                     & & \textbf{c}^{i}_{k,m,l} \in \mathcal{L}^{i,j}_{k,m,l} & & \forall j \in \mathcal{N}^{i}, m, l \\
&                     & & (\ref{eq: communication range1}), (\ref{eq: communication range2}), (\ref{eq: equality constraints}), (\ref{eq: dynamic feasible constraints})
\end{aligned}
\label{eq: trajectory optimization}
\end{equation}

\section{Theoretical Guarantee}
\label{sec: theoretical guarantee}
In this section, we present the theoretical guarantee of the proposed algorithm.

\begin{lemma}
(Existence and safety of SFC) Assume that $\textbf{c}^{i}_{k-1,m,l} \in \mathcal{S}^{i}_{k-1,m}$ for $\forall m,l$ at the replanning step $k>0$. Then, there exists a non-empty convex set $\mathcal{S}^{i}_{k,m}$ that satisfies (\ref{eq: safe flight corridor construction}) and  $(\mathcal{S}^{i}_{k,m} \oplus \mathcal{C}^{i,o}) \cap \mathcal{O} = \emptyset$ for $\forall k, m$.
\label{lemma: safety of safe flight corridor}
\end{lemma}

\begin{proof}
If $k=0$, $\mathcal{S}^{i}_{0,m} = \langle \textbf{s}^{i}, \textbf{w}^{i}_{0} \rangle$ satisfies (\ref{eq: safe flight corridor construction}) and $(\mathcal{S}^{i}_{k,m} \oplus \mathcal{C}^{i,o}) \cap \mathcal{O} = \emptyset$ since the agent does not collide with static obstacles when it is on the grid.

Assume that $\mathcal{S}^{i}_{k-1,m}$ satisfies (\ref{eq: safe flight corridor construction}) and $(\mathcal{S}^{i}_{k-1,m} \oplus \mathcal{C}^{i,o}) \cap \mathcal{O} = \emptyset$. 
If $m<M$, then $\mathcal{S}^{i}_{k,m} = \mathcal{S}^{i}_{k-1,m}$ satisfies (\ref{eq: safe flight corridor construction}) and $(\mathcal{S}^{i}_{k,m} \oplus \mathcal{C}^{i,o}) \cap \mathcal{O} = \emptyset$ by the assumption.
If $m=M$ and (\ref{eq: safe flight corridor condition}) is satisfied, then $\mathcal{S}^{i}_{k,M} = \text{Conv}(\{\hat{\textbf{c}}^{i}_{k,M,n},\textbf{g}^{i}_{k-1},\textbf{w}^{i}_{k}\})$ satisfies (\ref{eq: safe flight corridor construction}) and $(\mathcal{S}^{i}_{k,M} \oplus \mathcal{C}^{i,o}) \cap \mathcal{O} = \emptyset$ by (\ref{eq: safe flight corridor condition}).
If $m=M$ and (\ref{eq: safe flight corridor condition}) is not satisfied, we can construct the SFC as $\mathcal{S}^{i}_{k,M} = \text{Conv}(\{\hat{\textbf{c}}^{i}_{k,M,n},\textbf{g}^{i}_{k-1}\})$.
It satisfies (\ref{eq: safe flight corridor construction}) and $\mathcal{S}^{i}_{k,m} = \text{Conv}(\{\hat{\textbf{c}}^{i}_{k,M,n},\textbf{g}^{i}_{k-1}\}) = \text{Conv}(\{\textbf{c}^{i}_{k-1,M,n},\textbf{g}^{i}_{k-1}\}) \subset \mathcal{S}^{i}_{k-1,M}$, which implies $(\mathcal{S}^{i}_{k,m} \oplus \mathcal{C}^{i,o}) \cap \mathcal{O} = \emptyset$ by the assumption.
Thus, there exists $\mathcal{S}^{i}_{k,m}$ that satisfies (\ref{eq: safe flight corridor construction}) and $(\mathcal{S}^{i}_{k,m} \oplus \mathcal{C}^{i,o}) \cap \mathcal{O} = \emptyset$ for every replanning step by mathematical induction.
\end{proof}

To prove the safety of LSC, we first linearize the inter-agent collision constraint (\ref{eq: definition of inter-agent avoidance}).
Let define the inter-agent collision model $\mathcal{C}^{i,j}$ as follows:
\begin{equation}
  \mathcal{C}^{i,j}=\{\textbf{x} \in \mathbb{R}^{2} \mid \|\textbf{x}\| < 2r\}
\label{eq: inter collision model}
\end{equation}
We define $\mathcal{H}^{i,j}_{k,m}$ as the convex hull of the control points of relative trajectory between the agents $i$ and $j$, i.e.:
\begin{equation}
    \mathcal{H}^{i,j}_{k,m} = \text{Conv}(\{\textbf{c}^{i}_{k,m,l} - \textbf{c}^{j}_{k,m,l} \mid l=0,\cdots,n\})
\label{eq: convex hull of control point of relative trajectory}
\end{equation}
Lemma $\ref{lemma: linearized inter-agent avoidance}$ shows the sufficient condition of (\ref{eq: definition of inter-agent avoidance}) using the same approach to \cite{park2022online}.

\begin{lemma}
(Inter-collision avoidance) If $\mathcal{H}^{i,j}_{k,m} \cap \mathcal{C}^{i,j} = \emptyset$ for $\forall m=1,\cdots,M$, then $\|\textbf{p}^{i}_{k}(t) - \textbf{p}^{j}_{k}(t)\| \geq 2r, \forall t \in [T_{k}, T_{k+M}]$.
\label{lemma: linearized inter-agent avoidance}
\end{lemma}

\begin{proof}
The relative trajectory between two agents is a piecewise Bernstein polynomial. The $m^{th}$ segment of the relative trajectory can be represented as follows:
\begin{equation}
  \begin{multlined}
    \textbf{p}^{i}_{k}(t) - \textbf{p}^{j}_{k}(t) = \sum_{l=0}^{n}(\textbf{c}^{i}_{k,m,l}-\textbf{c}^{j}_{k,m,l})b_{l,n}(\frac{t-T_{k+m-1}}{\Delta t}), \\ \forall t \in [T_{k+m-1, k+m}]
 \end{multlined}
\label{eq: linearized collision avoidance constraint proof1}
\end{equation}
Due to the convex hull property of the Bernstein polynomial, we obtain the following for $\forall m, t \in [T_{k+m-1}, T_{k+m}]$.
\begin{equation}
    \textbf{p}^{i}_{k}(t) - \textbf{p}^{j}_{k}(t) \in \mathcal{H}^{i,j}_{k,m}, \forall t \in [T_{k+m-1, k+m}]
\label{eq: linearized collision avoidance constraint proof2}
\end{equation}
\begin{equation}
    \textbf{p}^{i}_{k}(t) - \textbf{p}^{j}_{k}(t) \notin \mathcal{C}^{i,j}, \forall t \in [T_{k+m-1, k+m}]
\label{eq: linearized collision avoidance constraint proof3}
\end{equation}
\begin{equation}
    \|\textbf{p}^{i}_{k}(t) - \textbf{p}^{j}_{k}(t)\| \geq 2r, \forall t \in [T_{k+m-1, k+m}]
\label{eq: linearized collision avoidance constraint proof4}
\end{equation}
This concludes the proof.
\end{proof}

\begin{lemma}
(Safety of LSC) If $\textbf{c}^{i}_{k,m,l} \in \mathcal{L}^{i,j}_{k,m,l}$, $\textbf{c}^{j}_{k,m,l} \in \mathcal{L}^{j,i}_{k,m,l}$ for $\forall m,l$ then we obtain $\mathcal{H}^{i,j}_{k,m} \cap  \mathcal{C}^{i,j} = \emptyset$ which implies that the agent $i$ does not collide with the agent $j$.
\label{lemma: safety of linear safe corridor}
\end{lemma}

\begin{proof}
We obtain the following by adding the inequality in (\ref{eq: linear safe corridor}) for each agent $i$ and $j$:
\begin{equation}
    (\textbf{c}^{i}_{k,m,l}-\hat{\textbf{c}}^{j}_{k,m,l}) \cdot \textbf{n}^{i,j}_{m} + (\textbf{c}^{j}_{k,m,l}-\hat{\textbf{c}}^{i}_{k,m,l}) \cdot \textbf{n}^{j,i}_{m} - (d^{i,j}_{m,l} + d^{j,i}_{m,l}) \geq 0
\label{eq: linear safe corridor safety1}
\end{equation}
This can be simplified as follows using (\ref{eq: linear safe corridor}):
\begin{equation}
  (\textbf{c}^{i}_{k,m,l}-\textbf{c}^{j}_{k,m,l}) \cdot \textbf{n}^{i,j}_{m} + (\hat{\textbf{c}}^{i}_{k,m,l}-\hat{\textbf{c}}^{j}_{k,m,l}) \cdot \textbf{n}^{i,j}_{m} 
  - (d^{i,j}_{m,l} + d^{j,i}_{m,l}) \geq 0
\label{eq: linear safe corridor safety2}
\end{equation}
\begin{equation}
  (\textbf{c}^{i}_{k,m,l}-\textbf{c}^{j}_{k,m,l}) \cdot \textbf{n}^{i,j}_{m} \geq 2r
\label{eq: linear safe corridor safety3}
\end{equation}
The above inequality satisfies for $\forall l$, thus we have the following for any $\lambda_{\forall l} \geq 0$ s.t. $\sum_{l=0}^{n}\lambda_{l}=1$:
\begin{equation}
  \sum_{l=0}^{n}\lambda_{l}(\textbf{c}^{i}_{k,m,l}-\textbf{c}^{j}_{k,m,l}) \cdot \textbf{n}^{i,j}_{m} \geq 2r
\label{eq: linear safe corridor safety4}
\end{equation}
\begin{equation}
  \mathcal{H}^{i,j}_{k,m} \cap  \mathcal{C}^{i,j} = \emptyset
\label{eq: linear safe corridor safety5}
\end{equation}
Thus, there is no collision between the agents $i$ and $j$ by Lemma \ref{lemma: linearized inter-agent avoidance}.
\end{proof}

\begin{theorem}
(Collision avoidance) The trajectory from (\ref{eq: trajectory optimization}) does not cause inter-agent collision or collision between agent and obstacle.
\label{theorem: collision avoidance}
\end{theorem}

\begin{proof}
The agent does not collide with static obstacles due to Lemma \ref{lemma: safety of safe flight corridor}.
For the agent $j \in \mathcal{N}^{i}$, there is no collision between the agents $i$ and $j$ due to Lemma \ref{lemma: safety of linear safe corridor}.
For the agent $j \notin \mathcal{N}^{i}$, we have the following inequality for all agents due to (\ref{eq: communication range1}):
\begin{equation}
  \|\textbf{c}^{i}_{k,m,l} - \textbf{c}^{i}_{k,1,0}\|_{\infty} \leq \frac{r_{c}}{2} - r, \forall i \in \mathcal{I}, m, l
\label{eq: safety1}
\end{equation}
Due to the convex hull property and end-point property of Bernstein polynomial \cite{farouki2012bernstein}, we obtain the following for all agent $i \in \mathcal{I}$:
\begin{equation}
  \|\textbf{p}^{i}_{k}(t) - \textbf{c}^{i}_{k,1,0}\|_{\infty} \leq \frac{r_{c}}{2} - r, \forall i \in \mathcal{I}, t \in [T_{k}, T_{k+M}]
\label{eq: safety2}
\end{equation}
\begin{equation}
  \|\textbf{p}^{i}_{k}(t) -\textbf{p}^{i}_{k}(T_{k}) \|_{\infty} \leq \frac{r_{c}}{2} - r, \forall i \in \mathcal{I}, t \in [T_{k}, T_{k+M}]
\label{eq: safety3}
\end{equation}
Since we assume that $j \notin \mathcal{N}^{i}$:
\begin{equation}
  \|\textbf{p}^{i}_{k}(T_{k}) - \textbf{p}^{j}_{k}(T_{k})\|_{\infty} \geq r_{c}
\label{eq: safety4}
\end{equation}
\begin{equation}
  \|\textbf{p}^{i}_{k}(T_{k}) - \textbf{p}^{i}_{k}(t) +
  \textbf{p}^{j}_{k}(t) - \textbf{p}^{j}_{k}(T_{k}) + 
  \textbf{p}^{i}_{k}(t) - \textbf{p}^{j}_{k}(t)\|_{\infty} \geq r_{c}
\label{eq: safety5}
\end{equation}
According to (\ref{eq: safety3}) and triangle inequality:
\begin{equation}
\begin{aligned}
  &\|\textbf{p}^{i}_{k}(T_{k}) - \textbf{p}^{i}_{k}(t)\| + \|\textbf{p}^{j}_{k}(t) - \textbf{p}^{j}_{k}(T_{k})\| \\ 
  &+ \|\textbf{p}^{i}_{k}(t) - \textbf{p}^{j}_{k}(t)\|_{\infty} \geq r_{c}
\end{aligned}
\label{eq: safety6}
\end{equation}
\begin{equation}
    \|\textbf{p}^{i}_{k}(t) - \textbf{p}^{j}_{k}(t)\| \geq \|\textbf{p}^{i}_{k}(t) - \textbf{p}^{j}_{k}(t)\|_{\infty} \geq 2r
\label{eq: safety7}
\end{equation}
Therefore, there is no collision between the agents $i$ and $j$.
In conclusion, the trajectory from (\ref{eq: trajectory optimization}) does not cause collision.
\end{proof}

\begin{lemma}
(Feasibility of SFC) Assume that $\textbf{c}^{i}_{k-1,m,l} \in \mathcal{S}^{i}_{k-1,m}$ for $\forall m,l$ at the replanning step $k>0$. Then,  $\hat{\textbf{c}}^{i}_{k,m,l} \in \mathcal{S}^{i}_{k,m}$ for $\forall k, m, l$.
\label{lemma: feasibility of safe flight corridor}
\end{lemma}

\begin{proof}
If $k=0$, we obtain $\hat{\textbf{c}}^{i}_{0,m,l} = \textbf{s}^{i} \in \mathcal{S}^{i}_{0,m}$ for $\forall m, l$ due to (\ref{eq: control point of initial trajectory}).
If $k>0$ and $m<M$, we have $\hat{\textbf{c}}^{i}_{k,m,l} = \textbf{c}^{i}_{k-1,m+1,l} \in \mathcal{S}^{i}_{k-1,m+1} = \mathcal{S}^{i}_{k,m}$ for $\forall l$ due to (\ref{eq: control point of initial trajectory}) and (\ref{eq: safe flight corridor construction}).
If $k>0$ and $m=M$, we obtain $\hat{\textbf{c}}^{i}_{k,M,l} = \hat{\textbf{c}}^{i}_{k,M,n} \in \mathcal{S}(\{\hat{\textbf{c}}^{i}_{k,M,n},\textbf{g}^{i}_{k-1}\}) \subset \mathcal{S}^{i}_{k,M}$ for $\forall l$ by (\ref{eq: control point of initial trajectory}) and (\ref{eq: safe flight corridor construction}).
Thus, we obtain $\hat{\textbf{c}}^{i}_{k,m,l} \in \mathcal{S}^{i}_{k,m}$ for $\forall k,m,l$.
\end{proof}

To prove the feasibility of LSC, we define $\hat{\mathcal{H}}^{i,j}_{k,m}$ as the convex hull of the control points of relative initial trajectory between the agents $i$ and $j$, i.e.:
\begin{equation}
    \hat{\mathcal{H}}^{i,j}_{k,m} = \text{Conv}(\{\hat{\textbf{c}}^{i}_{k,m,l} - \hat{\textbf{c}}^{j}_{k,m,l} \mid l=0,\cdots,n\})
\label{eq: convex hull of control point of initial trajectory}
\end{equation}

\begin{lemma}
(Feasibility of LSC) 
Assume that $\hat{\textbf{c}}^{i}_{k-1,m,l} \in \mathcal{L}^{i,j}_{k-1,m,l}$ for $\forall j \in \mathcal{N}^{i}, m, l$ at the replanning step $k > 0$, then there exists $\mathcal{L}^{i,j}_{k,m,l}$ that satisfies (\ref{eq: linear safe corridor}), (\ref{eq: linear safe corridor last segment}), and $\hat{\textbf{c}}^{i}_{k,m,l} \in \mathcal{L}^{i,j}_{k,m,l}$ for $\forall j \in \mathcal{N}^{i}, k, m, l$.
\label{lemma: feasibility of linear safe corridor}
\end{lemma}

\begin{proof}
If $k = 0$, we can set the normal vector of LSC as follows:
\begin{equation}
    \textbf{n}^{i,j}_{m} = \frac{\textbf{s}^{i} - \textbf{s}^{j}}{\|\textbf{s}^{i} - \textbf{s}^{j}\|}
\label{eq: feasibility of lsc1}
\end{equation}
Therefore, we obtain $\hat{\textbf{c}}^{i}_{0,m,l} = \textbf{s}^{i} \in \mathcal{L}^{i,j}_{0,m}$ for $\forall j \in \mathcal{N}^{i}, m, l$:
\begin{equation}
\begin{multlined}
    \textbf{s}^{i} \in \mathcal{L}^{i,j}_{0,m} \Leftrightarrow (\textbf{s}^{i} - \hat{\textbf{c}}^{j}_{k,m,l}) \cdot \textbf{n}^{i,j}_{m} - d^{i,j}_{m,l} \geq 0 \\ 
    \Leftrightarrow (\textbf{s}^{i} - \textbf{s}^{j}) \cdot \textbf{n}^{i,j}_{m} - (r + \frac{1}{2}(\textbf{s}^{i} - \textbf{s}^{j}) \cdot \textbf{n}^{i,j}_{m} ) \geq 0 \\
    \Leftrightarrow \frac{1}{2}(\textbf{s}^{i} - \textbf{s}^{j}) \cdot \textbf{n}^{i,j}_{m} - r \geq 0
    \Rightarrow \|\textbf{s}^{i} - \textbf{s}^{j}\| \geq 2r 
\end{multlined}
\label{eq: feasibility of lsc2}
\end{equation}

If $k > 0$ and $m < M$, we have the following by (\ref{eq: control point of initial trajectory}):
\begin{equation}
  \hat{\mathcal{H}}^{i,j}_{k,m} = \mathcal{H}^{i,j}_{k-1,m+1}
\label{eq: feasibility of lsc3}
\end{equation}
Therefore, $\hat{\mathcal{H}}^{i,j}_{k,m}$ and $\mathcal{C}^{i,j}$ are disjoint non-empty convex sets due to Lemma \ref{lemma: feasibility of linear safe corridor}:
\begin{equation}
  \hat{\mathcal{H}}^{i,j}_{k,m} \cap \mathcal{C}^{i,j} = \emptyset
\label{eq: feasibility of lsc4}
\end{equation}
By the hyperplane separation theorem \cite{boyd2004convex}, there exists $\textbf{n}_{s}$ such that:
\begin{equation}
   \min \langle \hat{\mathcal{H}}^{i,j}_{k,m}, \textbf{n}_{s} \rangle \geq 2r
\label{eq: feasibility of lsc5}
\end{equation}
where $\min \langle \hat{\mathcal{H}}^{i,j}_{k,m}, \textbf{n}_{s} \rangle = \min_{\textbf{x} \in \hat{\mathcal{H}}^{i,j}_{k,m}} \textbf{x} \cdot \textbf{n}_{s}$.
Here, we set the normal vector of LSC as follows:
\begin{equation}
  \textbf{n}^{i,j}_{m} = -\textbf{n}^{j,i}_{m} = \textbf{n}_{s}
\label{eq: feasibility of lsc6}
\end{equation}
Then, we obtain the following:
\begin{equation}
\begin{multlined}
  (\hat{\textbf{c}}^{i}_{k,m,l} - \hat{\textbf{c}}^{j}_{k,m,l}) \cdot \textbf{n}^{i,j}_{m} - d^{i,j}_{m,l} \geq 0 \\
  \Leftrightarrow  (\hat{\textbf{c}}^{i}_{k,m,l} - \hat{\textbf{c}}^{j}_{k,m,l}) \cdot \textbf{n}^{i,j}_{m} - 2r \geq 0 \Leftarrow (\ref{eq: feasibility of lsc4})
\end{multlined}
\label{eq: feasibility of lsc7}
\end{equation}
Therefore, $\hat{\textbf{c}}^{i}_{k,m,l} \in \mathcal{L}^{i,j}_{k,m,l}$ for $\forall j \in \mathcal{N}^{i}, m < M, l$.

If $k > 0$ and $m = M$, then we have $\hat{\textbf{c}}^{i}_{k,M,l} = \hat{\textbf{c}}^{i}_{k,M,n}$ for $\forall l$ by (\ref{eq: control point of initial trajectory}).
Also, the LSC when $m = M$ satisfies $\hat{\textbf{c}}^{i}_{k,M,n} \in \mathcal{L}^{i,j}_{k,M,l}$ for $\forall l$ because $\langle \hat{\textbf{c}}^{i}_{k,M,n}, \textbf{g}^{i}_{k-1} \rangle \subset \mathcal{L}^{i,j}_{k,M,l}$ for $\forall l$. 
Therefore, $\hat{\textbf{c}}^{i}_{k,M,l} \in \mathcal{L}^{i,j}_{k,M,l}$ for $\forall j \in \mathcal{N}^{i}, l$.
This concludes the proof.
\end{proof}

\begin{lemma}
(Feasibility of subgoal optimization) The solution of (\ref{eq: subgoal optimization}) always exists for every replanning step.
\label{lemma: feasibility of subgoal optimization}
\end{lemma}

\begin{proof}
If $k = 0$, $\textbf{s}^{i}$ satisfies all the constraints of (\ref{eq: subgoal optimization}).
If $k > 0$, $\textbf{g}^{i}_{k-1}$ satisfies all the constraints because $\textbf{g}^{i}_{k-1} \in \mathcal{S}^{i}_{k,M}$ by (\ref{eq: safe flight corridor construction}) and $\textbf{g}^{i}_{k-1} \in \mathcal{L}^{i,j}_{k,M,n}$ by (\ref{eq: linear safe corridor last segment}).
This concludes the proof.
\end{proof}

\begin{theorem}
(Feasibility of optimization problem) If the replanning period is equal to the segment duration $\Delta t$, the solution of (\ref{eq: trajectory optimization}) always exists for every replanning step.
\label{theorem: feasibility of optimization problem}
\end{theorem}

\begin{proof}
We can prove this by showing that the $\hat{\textbf{p}}^{i}_{k}(t)$ satisfies all the constraints in (\ref{eq: trajectory optimization}) for every replanning step.

If $k = 0$, $\hat{\textbf{p}}^{i}_{k}(t)$ satisfies SFC and LSC constraints due to Lemmas \ref{lemma: feasibility of safe flight corridor}, \ref{lemma: feasibility of linear safe corridor}. $\hat{\textbf{p}}^{i}_{k}(t)$ fulfills the initial condition, continuity constraint, final stop condition, dynamical limit constraints, and (\ref{eq: communication range1}) because $\hat{\textbf{c}}^{i}_{0,m,l} = \textbf{s}^{i}$ for $\forall m,l$. $\hat{\textbf{p}}^{i}_{k}(t)$ satisfies (\ref{eq: communication range2}) due to the assumption that $d > 2\sqrt{2}r$.
Therefore, $\hat{\textbf{p}}^{i}_{k}(t)$ is one of the solution that satisfies all the constraints in (\ref{eq: trajectory optimization}).

If there exists a solution at the previous replanning step $k-1$, then $\hat{\textbf{p}}^{i}_{k}(t)$ satisfies SFC and LSC constraints due to Lemmas \ref{lemma: feasibility of safe flight corridor}, \ref{lemma: feasibility of linear safe corridor}.
$\hat{\textbf{p}}^{i}_{k}(t)$ fulfills the initial condition, continuity constraint, final stop condition, and dynamical limit constraints, (\ref{eq: communication range1}), and (\ref{eq: communication range2}) due to (\ref{eq: control point of initial trajectory}).
Therefore, the solution of (\ref{eq: trajectory optimization}) always exists for every replanning step by mathematical induction.
\end{proof}

\begin{lemma}
(Sufficient conditions of deadlock resolution)
If the agent $i$ satisfies the following, then it does not cause deadlock: 
\begin{equation}
  \textbf{c}^{i}_{k,M,n} \neq \textbf{g}^{i}_{k}
\label{eq: deadlock sufficient condition1}
\end{equation}
\begin{equation}
  \textbf{g}^{i}_{k} \in \mathcal{S}^{i}_{k,M},\;\; \textbf{g}^{i}_{k} \in \mathcal{L}^{i,\forall j \in \mathcal{N}^{i}}_{k,M,n},\;\; \|\textbf{g}^{i}_{k} - \textbf{w}^{i}_{k}\|_{\infty} \leq \frac{r_{c}}{2}
\label{eq: deadlock sufficient condition2}
\end{equation}
\label{lemma: sufficient conditions of deadlock resolution}
\end{lemma}

\begin{proof}
Assume that the agent $i$ causes deadlock at the $\textbf{p}_{0}$ when $k > k_{0}$, i.e.:
\begin{equation}
    \textbf{c}^{i}_{k,m,l} = \textbf{p}_{0} \neq \textbf{g}^{i}_{k}, \forall k > k_{0},m,l
\label{eq: deadlock sufficient condition3}
\end{equation}
It implies that $\textbf{p}^{i}_{k}(t) = \textbf{p}_{0}$ is the optimal solution of (\ref{eq: trajectory optimization}).
Here, we define another trajectory $\tilde{\textbf{p}}^{i}_{k}(t)$ that has the following control points:
\begin{equation}
    \begin{alignedat}{2}
    \tilde{\textbf{c}}^{i}_{k,m,l} = 
    \begin{cases} 
    \ \textbf{p}_{0} + \delta (\textbf{g}^{i}_{k} - \textbf{p}_{0})   & m = M, l \geq n-2 \\
    \ \textbf{p}_{0}   &  else \\    
    \end{cases}
\end{alignedat}
\label{eq: deadlock sufficient condition3-2}
\end{equation}
where $\tilde{\textbf{c}}^{i}_{k,m,l}$ is the control point of $\tilde{\textbf{p}}^{i}_{k}(t)$, and $\delta \in [0,1]$. It satisfies the initial condition, continuity constraint, and final stop condition. Let the feasible region of the collision constraints and (\ref{eq: communication range2}) for $\textbf{c}^{i}_{k,M,l}$ be $\mathcal{F}_{l}$.
Then, we obtain $\tilde{\textbf{c}}^{i}_{k,M,l} \in \mathcal{F}_{l}$ for $\forall l$ because $\mathcal{F}_{l}$ is a convex set and $\textbf{p}_{0} \in \mathcal{F}_{l}$, $\textbf{g}^{i}_{k} \in \mathcal{F}_{l}$ by (\ref{eq: deadlock sufficient condition2}).
It indicates that  $\tilde{\textbf{p}}^{i}_{k}(t)$ fulfills the collision constraints and (\ref{eq: communication range2}).
Therefore, $\tilde{\textbf{p}}^{i}_{k}(t)$ is a feasible solution of (\ref{eq: trajectory optimization}) if $\delta$ satisfies the following:
\begin{equation}
    \|\delta (\textbf{g}^{i}_{k} - \textbf{p}_{0})\|_{\infty} \leq \frac{r_{c}}{2} - r
\label{eq: deadlock sufficient condition4}
\end{equation}
\begin{equation}
    \left\| \frac{n\delta (\textbf{g}^{i}_{k} - \textbf{p}_{0})}{\Delta t}\right\|_{\infty} \leq v_{max}
\label{eq: deadlock sufficient condition5}
\end{equation}
\begin{equation}
    \|\frac{n(n-1)\delta (\textbf{g}^{i}_{k} - \textbf{p}_{0})}{\Delta t^{2}}\|_{\infty} \leq a_{max}
\label{eq: deadlock sufficient condition6}
\end{equation}
where (\ref{eq: deadlock sufficient condition4}) is the sufficient condition of the communication range constraint (\ref{eq: communication range1}), and (\ref{eq: deadlock sufficient condition5}), (\ref{eq: deadlock sufficient condition6}) are the sufficient conditions of dynamical limit constraints. 
Note that if $\delta$ is small enough, there exists non-zero $\delta$ that satisfies the above constraints.

The cost of $\tilde{\textbf{p}}^{i}_{k}(t)$ can be represented as follows:
\begin{equation}
    J(\tilde{\textbf{p}}^{i}_{k}(t)) = w_{err}\|(\textbf{g}^{i}_{k} - \textbf{p}_{0})\|^{2}(1-\delta)^{2} + C\delta^{2}
\label{eq: deadlock sufficient condition7}
\end{equation}
where $J(\tilde{\textbf{p}}^{i}_{k}(t))$ is the cost of the $\tilde{\textbf{p}}^{i}_{k}(t)$, and $C > 0$ is the constant that is derived from the jerk minimization cost (\ref{eq: objective function derivative of trajectory}) of $\tilde{\textbf{p}}^{i}_{k}(t)$. 
The cost difference between $\tilde{\textbf{p}}^{i}_{k}(t)$ and $\textbf{p}^{i}_{k}(t)$ is:
\begin{equation}
    \Delta J = w_{err}\|(\textbf{g}^{i}_{k} - \textbf{p}_{0})\|^{2} (1-(1-\delta)^{2}) - C\delta^{2}
\label{eq: deadlock sufficient condition8}
\end{equation}
where $\Delta J = J(\textbf{p}^{i}_{k}(t)) - J(\tilde{\textbf{p}}^{i}_{k}(t))$.
Let us define $A = w_{err}\|(\textbf{g}^{i}_{k} - \textbf{p}_{0})\|^{2} > 0$. Then, we can simplify the (\ref{eq: deadlock sufficient condition8}) as follows:
\begin{equation}
    \Delta J = 2A\delta - (A+C)\delta^{2}
\label{eq: deadlock sufficient condition9}
\end{equation}
Here, we can observe that $\Delta J > 0$ when $0 < \delta < \frac{2A}{A+C}$.
It implies that $\textbf{p}^{i}_{k}(t)$ cannot be the optimal solution of (\ref{eq: trajectory optimization}) because we can always find a feasible solution $\tilde{\textbf{p}}^{i}_{k}(t)$ that has a lower cost than $\textbf{p}^{i}_{k}(t)$ by setting the non-zero $\delta$ small enough. This concludes the proof.
\end{proof}

\begin{theorem}
(Deadlock resolution) If $d > 2\sqrt{2}r$ and the MAPP in section \ref{subsec: decentralized MAPP} does not cause deadlock, then Alg. \ref{alg: trajectory planning algorithm} does not cause deadlock.
\label{theorem: deadlock resolution}
\end{theorem}

\begin{figure}
    \centering
    \begin{subfigure}[t]{0.30\textwidth}
        \includegraphics[width=\textwidth]{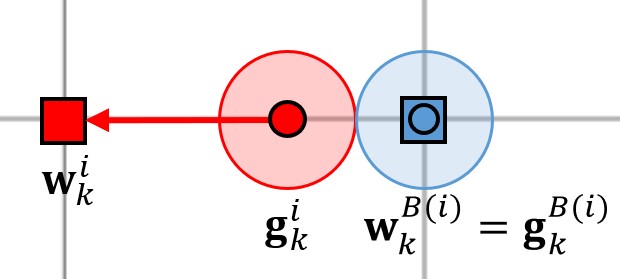}
        \caption{The position of agents when $B(i) \notin \mathcal{D}$ and $i \in \mathcal{D}$}
        \label{fig: proof1}
    \end{subfigure}
    ~ 
    \begin{subfigure}[t]{0.30\textwidth}
        \includegraphics[width=\textwidth]{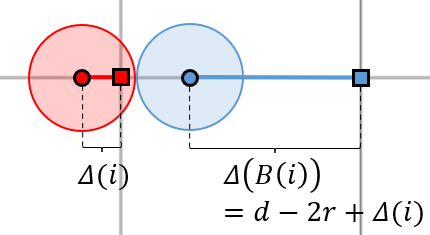}
        \caption{$\textbf{n}_{dir} > 0$}
        \label{fig: proof2}
    \end{subfigure}
    ~ 
    \begin{subfigure}[t]{0.30\textwidth}
        \includegraphics[width=\textwidth]{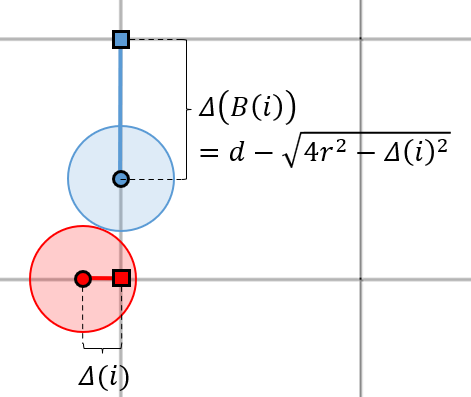}
        \caption{$\textbf{n}_{dir} \leq 0$}
        \label{fig: proof3}
    \end{subfigure}
    \caption{Illustrations for the proof of Theorem \ref{theorem: deadlock resolution}.
    The square dots are the waypoints and the circle dots are the subgoals. The circles denote the agent's current position.
    }
    \label{fig: proof for theorem 3}
    \vspace{-3mm}
\end{figure}

\begin{proof}
By Lemma \ref{lemma: sufficient conditions of deadlock resolution}, the agent does not cause deadlock if it satisfies (\ref{eq: deadlock sufficient condition1}) and (\ref{eq: deadlock sufficient condition2}).
Here, $\textbf{g}^{i}_{k}$ always satisfies (\ref{eq: deadlock sufficient condition2}) due to the constraints in (\ref{eq: subgoal optimization}) and the assumption that $r_{c} > 2d$. Therefore, Alg. \ref{alg: trajectory planning algorithm} causes deadlock if and only if there exists the agent $i$ such that:
\begin{equation}
  \textbf{c}^{i}_{k,M,n} = \textbf{g}^{i}_{k} = \textbf{g}^{i}_{k_{0}} \neq \textbf{w}^{i}_{k_{0}}, \forall k > k_{0}
\label{eq: deadlock resolution1}
\end{equation}
where $k_{0}$ is the replanning step when the deadlock happens.
Therefore, we can prove the theorem by showing that the subgoal optimization in Alg. \ref{alg: trajectory planning algorithm} prevents the deadlock condition (\ref{eq: deadlock resolution1}) for every replanning step.

The subgoal optimization problem (\ref{eq: subgoal optimization}) can be reformulated as follows:
\begin{equation}
\begin{aligned}
& \text{minimize}     & & \delta \\
& \text{subject to}   & & \delta \in [0,1] \\
&                     & & \textbf{w}^{i}_{k} + \delta (\textbf{g}^{i}_{k-1} - \textbf{w}^{i}_{k}) \in \mathcal{S}^{i}_{k,M} \\
&                     & & \textbf{w}^{i}_{k} + \delta (\textbf{g}^{i}_{k-1} - \textbf{w}^{i}_{k}) \in \mathcal{L}^{i,j}_{k,M,n}  & & \forall j \in \mathcal{N}^{i} \\
\end{aligned}
\label{eq: deadlock resolution2}
\end{equation}
where $\delta$ is the variable such that $\textbf{g}^{i}_{k} = \textbf{w}^{i}_{k} + \delta (\textbf{g}^{i}_{k-1} - \textbf{w}^{i}_{k})$. $\mathcal{S}^{i}_{k,M}$ can be represented as the intersection of linear constraints $a^{i}_{s} \delta - b^{i}_{s} \leq 0$, thus the Lagrangian function of (\ref{eq: deadlock resolution2}) is given as follows:
\begin{equation}
\begin{aligned}
L^{i} &= \delta - \lambda_{0} \delta + \lambda_{1} (\delta - 1) \\
&+ \sum_{s} \lambda^{i}_{s} (a^{i}_{s} \delta - b^{i}_{s}) \\
&+ \sum_{j \in \mathcal{N}^{i}} \lambda^{i,j} (d^{i,j}_{M,l} - (\textbf{w}^{i}_{k} + \delta (\textbf{g}^{i}_{k-1} - \textbf{w}^{i}_{k}) - \textbf{p}^{j}_{cls,i}) \cdot \textbf{n}^{i,j}_{M})
\end{aligned}
\label{eq: deadlock resolution3}
\end{equation}
where $\lambda_{0}$, $\lambda_{1}$, $\lambda^{i}_{s}$, and $\lambda^{i,j}$ are the Lagrangian multipliers.
Assume that there exists a set of agents $\mathcal{D} \subset \mathcal{I}$ that satisfies the deadlock condition (\ref{eq: deadlock resolution1}).
Then, if $k>k_{0}$, $\delta^{*} = 1$ is the optimal solution of the subgoal optimization problem for the agent $i \in \mathcal{D}$ due to (\ref{eq: deadlock resolution1}). Also, the agent always satisfies the SFC constraint regardless of $\delta \in [0,1]$ because $\langle \textbf{g}^{i}_{k-1}, \textbf{w}^{i}_{k} \rangle \subset \mathcal{S}^{i}_{k,M}$ by (\ref{eq: safe flight corridor construction}).
Thus, we obtain $\lambda_{0} = 0$ and $\lambda^{i}_{\forall s} = 0$ by the complementary slackness condition of KKT conditions \cite{boyd2004convex}.
Moreover, we have:
\begin{equation}
\frac{\partial L^{i}}{\partial \delta} = 1 + \lambda_{1} - \sum_{j \in \mathcal{N}^{i}} \lambda^{i,j} (\textbf{g}^{i}_{k-1} - \textbf{w}^{i}_{k}) \cdot \textbf{n}^{i,j}_{M} = 0
\label{eq: deadlock resolution4}
\end{equation}
due to the stationary condition of KKT conditions \cite{boyd2004convex}.
Since all agents remain static after deadlock, we have $\textbf{p}^{j}_{cls,i} = \textbf{g}^{j}_{k}$, $\textbf{p}^{i}_{cls,j} = \textbf{g}^{i}_{k}$ and $\textbf{g}^{i}_{k-1} = \textbf{g}^{i}_{k}$.
Using this, the above equation can be simplifed as follows:
\begin{equation}
\frac{\partial L^{i}}{\partial \delta} = 1 + \lambda_{1} - \sum_{j \in \mathcal{N}^{i}} \lambda^{i,j} \frac{(\textbf{g}^{i}_{k} - \textbf{w}^{i}_{k})^{T} (\textbf{g}^{i}_{k} - \textbf{g}^{j}_{k})}{\|\textbf{g}^{i}_{k} - \textbf{g}^{j}_{k}\|} = 0
\label{eq: deadlock resolution5}
\end{equation}
\begin{equation}
\sum_{j \in \mathcal{N}^{i}} \lambda^{i,j} \frac{(\textbf{g}^{j}_{k} - \textbf{g}^{i}_{k})^{T}(\textbf{w}^{i}_{k} - \textbf{g}^{i}_{k})}{\|\textbf{g}^{i}_{k} - \textbf{g}^{j}_{k}\|} = 1 + \lambda_{1}
\label{eq: deadlock resolution5_2}
\end{equation}
Since $\lambda_{1} \geq 0$ and $\lambda^{i,j} \geq 0$ by the dual feasibility of KKT conditions, there must exists the agent $j \in \mathcal{N}^{i}$ that satisfies $\lambda^{i,j} > 0$ and the following condition to fulfill (\ref{eq: deadlock resolution5_2}):
\begin{equation}
  (\textbf{g}^{j}_{k} - \textbf{g}^{i}_{k})^{T}(\textbf{w}^{i}_{k} - \textbf{g}^{i}_{k}) > 0, \forall k > k_{0}
\label{eq: deadlock resolution6}
\end{equation}
Moreover, the agent $j$ must satisfy the following for $\forall k > k_{0}$ by the complementary slackness of KKT conditions:
\begin{equation}
  d^{i,j}_{M,l} - (\textbf{w}^{i}_{k} + (\textbf{g}^{i}_{k-1} - \textbf{w}^{i}_{k}) - \textbf{p}^{j}_{cls,i}) \cdot \textbf{n}^{i,j}_{M} = 0
\label{eq: deadlock resolution7}
\end{equation}
\begin{equation}
  r + \frac{1}{2}\|\textbf{g}^{i}_{k} - \textbf{g}^{j}_{k}\| - (\textbf{g}^{i}_{k-1} - \textbf{p}^{j}_{cls,i}) \cdot \frac{\textbf{g}^{i}_{k} - \textbf{g}^{j}_{k}}{\|\textbf{g}^{i}_{k} - \textbf{g}^{j}_{k}\|} = 0
\label{eq: deadlock resolution8}
\end{equation}
\begin{equation}
  \|\textbf{g}^{i}_{k} - \textbf{g}^{j}_{k}\| = 2r, \forall k > k_{0}
\label{eq: deadlock resolution9}
\end{equation}
Let us define the agent $B(i) \in \mathcal{I}$ that satisfies (\ref{eq: deadlock resolution6}) and (\ref{eq: deadlock resolution9}) to a \textit{blocking agent} of the agent $i$, where $B(\cdot)$ indicates the blocking agent of the input. 
For a deadlock to occur, the agents in $\mathcal{D}$ must have their blocking agent.

Suppose that $B(i) \notin \mathcal{D}$ when $i \in \mathcal{D}$. Then, the agents $i$ and $B(i)$ must be on the same grid edge after the deadlock happens, as shown in Fig. \ref{fig: proof1}. It is because $B(i)$ converges to its waypoint and the distance between two agents is $2r$ by (\ref{eq: deadlock resolution9}). However, the waypoints of two agents must be different by Lemma \ref{lemma: no duplicated waypoints}, so $B(i)$ cannot satisfy (\ref{eq: deadlock resolution6}).
Thus, $B(i) \in \mathcal{D}$ for $\forall i \in \mathcal{D}$.

Let us choose an agent $i$ in $\mathcal{D}$ that satisfies the following:
\begin{equation}
  \Delta(i) \geq \Delta(j), \forall j \in \mathcal{D}
\label{eq: deadlock resolution10}
\end{equation}
where $\Delta(i) = \|\textbf{w}^{i}_{k} - \textbf{g}^{i}_{k}\|$.
As discussed earlier, the agent $i$ has its blocking agent $B(i) \in \mathcal{D}$, and the agents $i$ and $B(i)$ are on different grid edges due to (iii) of Lemma \ref{lemma: one agent per edge}.
Therefore, $\Delta(B(i))$ is computed as follows:
\begin{equation}
\begin{alignedat}{2}
    \Delta(B(i)) = 
    \begin{cases} 
    \ d - 2r + \Delta(i), &\text{if } \textbf{n}_{dir} > 0  \\
    \ d - \sqrt{4r^{2} - \Delta(i)^{2}}, &\text{else} \\
    \end{cases}
\end{alignedat}
\label{eq: deadlock resolution11}
\end{equation}
\begin{equation}
  \textbf{n}_{dir} = (\textbf{w}^{B(i)}_{k} - \textbf{g}^{B(i)}_{k})^{T}(\textbf{w}^{i}_{k} - \textbf{g}^{i}_{k})
\label{eq: deadlock resolution12}
\end{equation}
Fig. \ref{fig: proof2} and \ref{fig: proof3} illustrate the derivation process of the above equations.
Here, we can observe that $\Delta(B(i)) > \Delta(i)$ because $d > 2\sqrt{2}r$.
However, it is inconsistent with the assumption (\ref{eq: deadlock resolution10}).
Therefore, there is at least one agent that does not have a blocking agent.
Thus, Alg. \ref{alg: trajectory planning algorithm} does not cause deadlock.
\end{proof}


\section{EVALUATION}
\label{sec: evaluation}
We compared the following algorithms to verify the performance of the proposed algorithm:
\begin{itemize}
  \item LSC-PB (LSC with priority-based goal planning, \cite{park2022online})
  \item LSC-DR (LSC with deadlock resolution, \textbf{proposed})
\end{itemize}
We modeled the agent with radius $r = 0.15$ m, maximum velocity $v_{max} = 1.0$ m/s, maximum acceleration $a_{max} = 2.0$ m/$\text{s}^2$ based on the experimental result with Crazyflie 2.1.
To represent the trajectory, we set the degree of polynomials $n=5$, the number of segments $M=10$, and the segment time $\Delta t = 0.2$ s. Therefore, the total planning horizon is 2 s.
We assigned the replanning period to be $\Delta t = 0.2$ s to satisfy the assumption in Thm. \ref{theorem: feasibility of optimization problem}, so the trajectories are updated with the rate of 5 Hz at the same time.
For decentralized MAPP, we implemented PIBT based on the source code of \cite{okumura2021iterative}, and we set the grid size $d = 0.5$ m to fulfill the assumption that $d > 2\sqrt{2}r$. 
We used the Octomap library \cite{hornung2013octomap} to represent the obstacles, and we utilized randomized Prim's algorithm \cite{foltin2011automated} to generate mazes.
We set $w_{err} = 1$, $w_{der}=0.01$ as the parameters of the objective function, and we used the CPLEX solver \cite{cplex201612} for subgoal and trajectory optimization.
The simulation was executed on a laptop with Intel Core i7-9750H @ 2.60GHz CPU and 32G RAM. 

\subsection{Simulation}
\begin{figure}
    \centering
    \begin{subfigure}[t]{0.45\textwidth}
        \includegraphics[width=\textwidth]{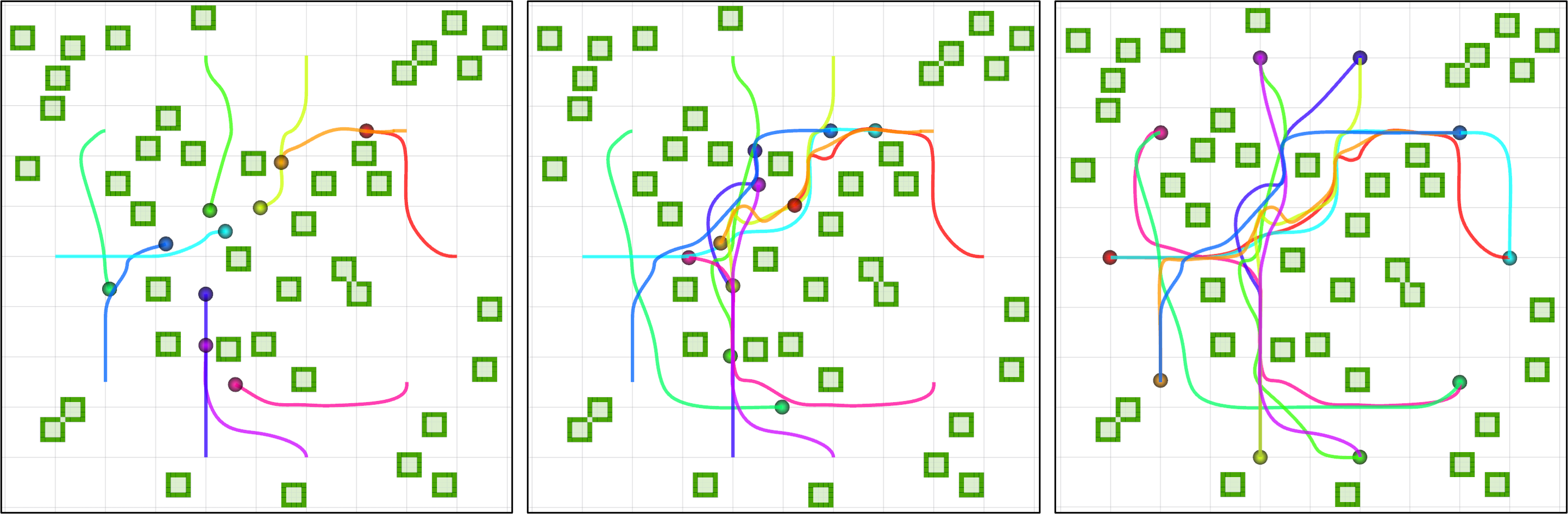}
        \caption{Random forest}
        \label{fig: random forest}
    \end{subfigure}
    ~ 
    \begin{subfigure}[t]{0.45\textwidth}
        \includegraphics[width=\textwidth]{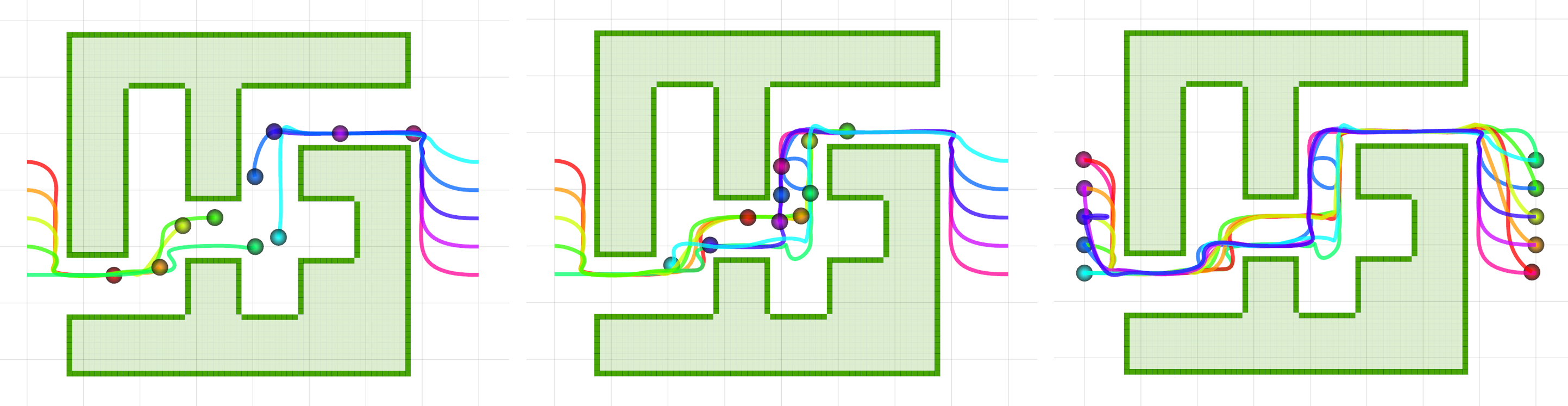}
        \caption{Sparse maze}
        \label{fig: sparse maze}
    \end{subfigure}
    ~ 
    \begin{subfigure}[t]{0.45\textwidth}
        \includegraphics[width=\textwidth]{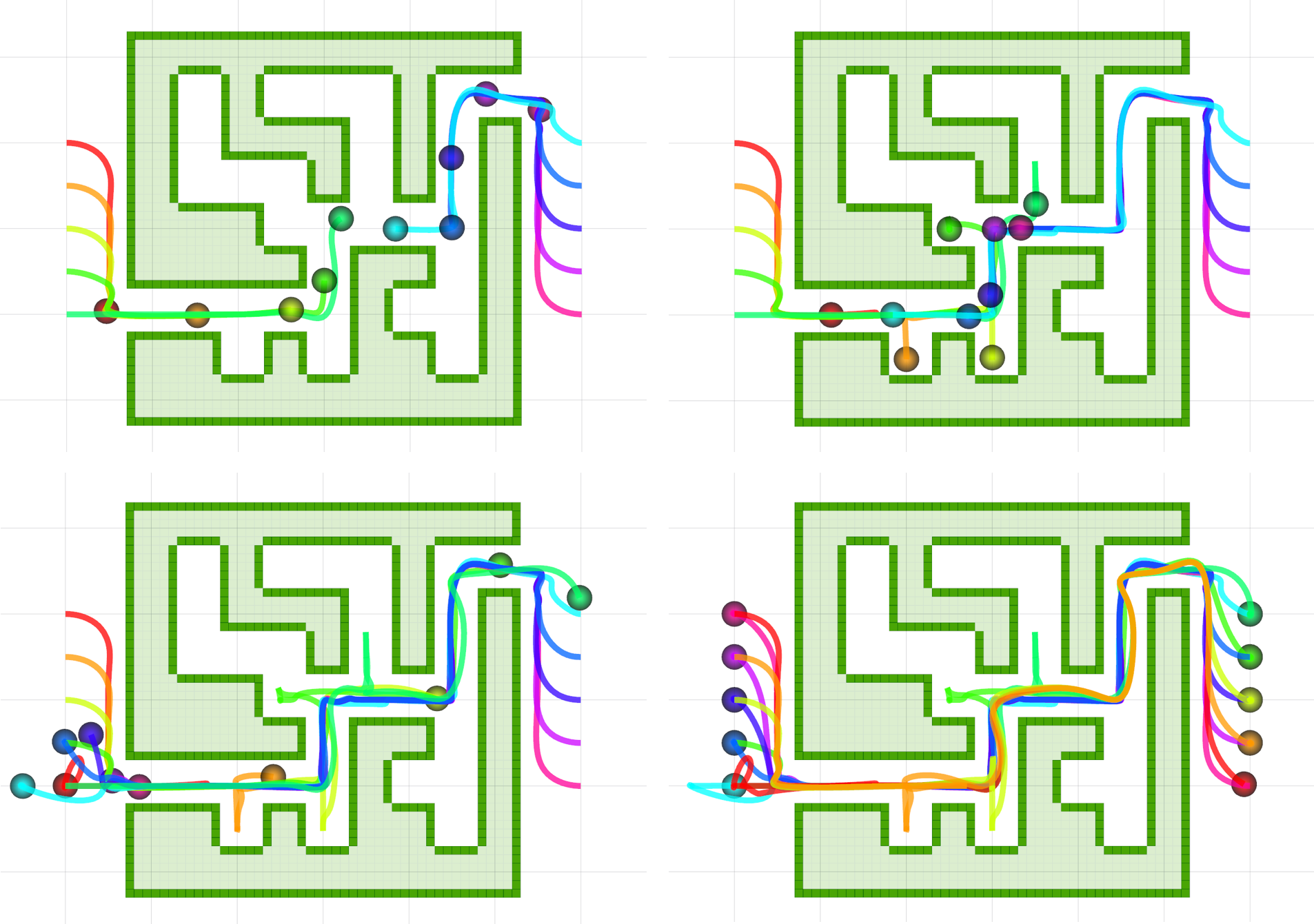}
        \caption{Dense maze}
        \label{fig: dense maze}
    \end{subfigure}
    \caption{Snapshot of trajectory generation by the proposed method ($r_{c} = 3$ m). The circle and line are the agent at its final location and its trajectory respectively, and the green-shaded region is the obstacle.
    }
    \label{fig: simulation result}
    \vspace{-3mm}
\end{figure}

We conducted the comparison in the following obstacle environments:
\begin{enumerate}[(i)]
  \item Random forest. We deploy 40 static obstacles in a random position and ten agents in a circle with 4 m radius. The goal point of the agent is at the antipodal point of the start point, as shown in Fig. \ref{fig: random forest}.
  \item Sparse maze. It consists of 6 $\times$ 6 cells, and each cell size is 1.0 m $\times$ 1.0 m, thus three agents can pass the corridor simultaneously. The maze has two entrances, and there are five agents at each entrance. We assigned each agent's goal point to the entrance on the other side of the maze, as depicted in Fig. \ref{fig: sparse maze}.
  \item Dense maze. It consists of 9 $\times$ 9 cells, and each cell size is 0.5 m $\times$ 0.5 m, thus only one agent can pass the corridor. We assigned the mission similar to the sparse maze, as illustrated in Fig. \ref{fig: dense maze}. 
\end{enumerate}
We judge that the mission failed when a collision occurred or when the agent failed to reach the goal within 60 s. For each map, we ran 30 trials changing the obstacle's position. 

Table \ref{table: cluttered environment} and Fig. \ref{fig: simulation result} describe the simulation results in obstacle environments. 
LSC-PB shows the perfect success rate in sparse environments and does not cause a collision in all cases.
However, LSC-PB fails to reach the goal in the dense maze because it cannot solve a deadlock when there is no space to yield to a higher priority agent.
On the other hand, our algorithm achieves the perfect success rate for all types of environments regardless of the communication range. 
It validates that the proposed algorithm can solve a deadlock even in a dense maze-like environment without a centralized coordinator.

The proposed method shows a 27.6$\%$ shorter flight time and a 7.4$\%$ shorter flight distance compared to LSC-PB when $r_{c} = \infty$. It is because LSC-PB performs a deadlock resolution only when the distance between the agents is close enough. Conversely, the proposed algorithm utilizes the final trajectory point, not the current position, for deadlock resolution. Therefore, the agent does not need to wait until other agents clump together.

\begin{table}[t]
\caption{Comparison with previous work \cite{park2022online}. The bold number indicates the best result ($sr$: success rate ($\%$), $T_{f}$: flight time (s), $L$: flight distance per agent (m), $T_{c}$: computation time (ms)).} \vspace{-3mm}
\label{table: cluttered environment}
\begin{center}
\begin{tabularx}{\linewidth}{c|c|*{4}{X}}
\toprule
Env. & Method & \centering $sr$ & \centering $T_{f}$ & \centering $L$ & \centering $T_{c}$ \tabularnewline
\hline 
\multirow{5}{*}{\makecell{Random \\ forest}} & LSC-PB \cite{park2022online} ($r_{c} = \infty$)  & \centering 100 &  \centering 25.7  & \centering 11.8 & \centering 8.15  \tabularnewline
& LSC-DR ($r_{c} = 2$ m) & \centering 100 & \centering 28.8 & \centering 11.7 & \centering \textbf{7.86}  \tabularnewline
& LSC-DR ($r_{c} = 3$ m) & \centering 100 & \centering 20.7 & \centering 11.3 & \centering 8.01  \tabularnewline
& LSC-DR ($r_{c} = 4$ m) & \centering 100 & \centering 19.9 & \centering 11.3 & \centering 8.23  \tabularnewline
& LSC-DR ($r_{c} = \infty$) & \centering 100 & \centering \textbf{19.1} & \centering \textbf{11.1} & \centering 8.16  \tabularnewline
\hline  
\multirow{5}{*}{\makecell{Sparse \\ maze}} & LSC-PB \cite{park2022online} ($r_{c} = \infty$)  & \centering 100 &  \centering 33.7  & \centering 13.9 & \centering 9.05  \tabularnewline
& LSC-DR ($r_{c} = 2$ m) & \centering 100 & \centering 34.4 & \centering 13.5 & \centering \textbf{8.51}  \tabularnewline
& LSC-DR ($r_{c} = 3$ m) & \centering 100 & \centering 27.1 & \centering 13.1 & \centering 8.62  \tabularnewline
& LSC-DR ($r_{c} = 4$ m) & \centering 100 & \centering \textbf{23.7} & \centering \textbf{12.6} & \centering 8.66  \tabularnewline
& LSC-DR ($r_{c} = \infty$) & \centering 100 & \centering 23.9 & \centering 12.7 & \centering 8.67  \tabularnewline
\hline  
\multirow{5}{*}{\makecell{Dense \\ maze}} & LSC-PB \cite{park2022online} ($r_{c} = \infty$)  & \centering 0 &  \centering -  & \centering - & \centering - \tabularnewline
& LSC-DR ($r_{c} = 2$ m) & \centering 100 & \centering 61.4 & \centering \textbf{16.5} & \centering 7.30  \tabularnewline
& LSC-DR ($r_{c} = 3$ m) & \centering 100 & \centering 51.0 & \centering 16.6 & \centering 7.44  \tabularnewline
& LSC-DR ($r_{c} = 4$ m) & \centering 100 & \centering 50.9 & \centering 17.1 & \centering 7.31  \tabularnewline
& LSC-DR ($r_{c} = \infty$) & \centering 100 & \centering \textbf{48.3} & \centering 16.7 & \centering \textbf{7.24} \tabularnewline
\hline
\end{tabularx}
\end{center}
\vspace{-5mm}
\end{table}

\subsection{Hardware demonstration}
Fig. \ref{fig: thumbnail} shows the hardware demonstration with ten Crazyflie 2.1 quadrotors in the dense maze. We set the communication range as 2 m, and we used the Crazyswarm \cite{preiss2017crazyswarm} to broadcast the trajectory to the agents. We present the full demonstration in the supplemental video, and there was no collision or deadlock during the entire flight. 

\section{CONCLUSIONS}
\label{sec: conclusions}
We presented the online decentralized MATP algorithm that guarantees to generate a safe, deadlock-free in a cluttered environment.
We utilized the decentralized MAPP for deadlock resolution, and generate the constraints so that the agent could reach the waypoint without deadlock. We proved that the proposed algorithm guarantees the feasibility of the optimization problem, collision avoidance, and deadlock-free for every replanning step. We verified that the proposed method does not cause collision or deadlock, regardless of the size of the free space or communication range.
Moreover, the proposed algorithm has a 27.6$\%$ shorter flight time and a 7.4$\%$ shorter flight distance than our previous work.
In future work, we plan to extend this algorithm to three-dimensional spaces, and we will try other MAPP algorithms to improve the performance further.

\newpage
\addtolength{\textheight}{-12cm}   









\bibliographystyle{./bibtex/IEEEtran}
\bibliography{./bibtex/IEEEabrv, ./bibtex/mybibfile}

\begin{thebibliography}{10}
\providecommand{\url}[1]{#1}
\csname url@rmstyle\endcsname
\providecommand{\newblock}{\relax}
\providecommand{\bibinfo}[2]{#2}
\providecommand\BIBentrySTDinterwordspacing{\spaceskip=0pt\relax}
\providecommand\BIBentryALTinterwordstretchfactor{4}
\providecommand\BIBentryALTinterwordspacing{\spaceskip=\fontdimen2\font plus
\BIBentryALTinterwordstretchfactor\fontdimen3\font minus
  \fontdimen4\font\relax}
\providecommand\BIBforeignlanguage[2]{{%
\expandafter\ifx\csname l@#1\endcsname\relax
\typeout{** WARNING: IEEEtran.bst: No hyphenation pattern has been}%
\typeout{** loaded for the language `#1'. Using the pattern for}%
\typeout{** the default language instead.}%
\else
\language=\csname l@#1\endcsname
\fi
#2}}

\bibitem{zhou2017fast}
D.~Zhou, Z.~Wang, S.~Bandyopadhyay, and M.~Schwager, ``Fast, on-line collision
  avoidance for dynamic vehicles using buffered voronoi cells,'' \emph{IEEE
  Robotics and Automation Letters}, vol.~2, no.~2, pp. 1047--1054, 2017.

\bibitem{luis2020online}
C.~E. {Luis}, M.~{Vukosavljev}, and A.~P. {Schoellig}, ``Online trajectory
  generation with distributed model predictive control for multi-robot motion
  planning,'' \emph{IEEE Robotics and Automation Letters}, vol.~5, no.~2, pp.
  604--611, 2020.

\bibitem{chen2022recursive}
Y.~Chen, M.~Guo, and Z.~Li, ``Recursive feasibility and deadlock resolution in
  mpc-based multi-robot trajectory generation,'' \emph{arXiv preprint
  arXiv:2202.06071}, 2022.

\bibitem{zhou2021ego}
X.~Zhou, J.~Zhu, H.~Zhou, C.~Xu, and F.~Gao, ``Ego-swarm: A fully autonomous
  and decentralized quadrotor swarm system in cluttered environments,'' in
  \emph{2021 IEEE International Conference on Robotics and Automation
  (ICRA)}.\hskip 1em plus 0.5em minus 0.4em\relax IEEE, 2021, pp. 4101--4107.

\bibitem{tordesillas2021mader}
J.~Tordesillas and J.~P. How, ``Mader: Trajectory planner in multiagent and
  dynamic environments,'' \emph{IEEE Transactions on Robotics}, 2021.

\bibitem{toumieh2022decentralized}
C.~Toumieh and A.~Lambert, ``Decentralized multi-agent planning using model
  predictive control and time-aware safe corridors,'' \emph{IEEE Robotics and
  Automation Letters}, 2022.

\bibitem{park2022online}
J.~Park, D.~Kim, G.~C. Kim, D.~Oh, and H.~J. Kim, ``Online distributed
  trajectory planning for quadrotor swarm with feasibility guarantee using
  linear safe corridor,'' \emph{IEEE Robotics and Automation Letters}, vol.~7,
  no.~2, pp. 4869--4876, 2022.

\bibitem{honig2018trajectory}
W.~H{\"o}nig, J.~A. Preiss, T.~S. Kumar, G.~S. Sukhatme, and N.~Ayanian,
  ``Trajectory planning for quadrotor swarms,'' \emph{IEEE Transactions on
  Robotics}, vol.~34, no.~4, pp. 856--869, 2018.

\bibitem{park2020efficient}
J.~{Park}, J.~{Kim}, I.~{Jang}, and H.~J. {Kim}, ``Efficient multi-agent
  trajectory planning with feasibility guarantee using relative bernstein
  polynomial,'' in \emph{2020 IEEE International Conference on Robotics and
  Automation (ICRA)}, 2020, pp. 434--440.

\bibitem{barer2014suboptimal}
M.~Barer, G.~Sharon, R.~Stern, and A.~Felner, ``Suboptimal variants of the
  conflict-based search algorithm for the multi-agent pathfinding problem,'' in
  \emph{Seventh Annual Symposium on Combinatorial Search}, 2014.

\bibitem{van2011reciprocal}
J.~Van Den~Berg, S.~J. Guy, M.~Lin, and D.~Manocha, ``Reciprocal n-body
  collision avoidance,'' in \emph{Robotics research}.\hskip 1em plus 0.5em
  minus 0.4em\relax Springer, 2011, pp. 3--19.

\bibitem{wang2017safety}
L.~Wang, A.~D. Ames, and M.~Egerstedt, ``Safety barrier certificates for
  collisions-free multirobot systems,'' \emph{IEEE Transactions on Robotics},
  vol.~33, no.~3, pp. 661--674, 2017.

\bibitem{zhu2019b}
H.~Zhu and J.~Alonso-Mora, ``B-uavc: Buffered uncertainty-aware voronoi cells
  for probabilistic multi-robot collision avoidance,'' in \emph{2019
  International Symposium on Multi-Robot and Multi-Agent Systems (MRS)}.\hskip
  1em plus 0.5em minus 0.4em\relax IEEE, 2019, pp. 162--168.

\bibitem{abdullhak2021deadlock}
M.~Abdullhak and A.~Vardy, ``Deadlock prediction and recovery for distributed
  collision avoidance with buffered voronoi cells,'' in \emph{2021 IEEE/RSJ
  International Conference on Intelligent Robots and Systems (IROS)}.\hskip 1em
  plus 0.5em minus 0.4em\relax IEEE, 2021, pp. 429--436.

\bibitem{jager2001decentralized}
M.~Jager and B.~Nebel, ``Decentralized collision avoidance, deadlock detection,
  and deadlock resolution for multiple mobile robots,'' in \emph{IEEE/RSJ
  International Conference on Intelligent Robots and Systems.}, vol.~3.\hskip
  1em plus 0.5em minus 0.4em\relax IEEE, 2001, pp. 1213--1219.

\bibitem{desaraju2012decentralized}
V.~R. Desaraju and J.~P. How, ``Decentralized path planning for multi-agent
  teams with complex constraints,'' \emph{Autonomous Robots}, vol.~32, no.~4,
  pp. 385--403, 2012.

\bibitem{alonso2018reactive}
J.~Alonso-Mora, J.~A. DeCastro, V.~Raman, D.~Rus, and H.~Kress-Gazit,
  ``Reactive mission and motion planning with deadlock resolution avoiding
  dynamic obstacles,'' \emph{Autonomous Robots}, vol.~42, no.~4, pp. 801--824,
  2018.

\bibitem{semnani2020force}
S.~H. Semnani, A.~H. de~Ruiter, and H.~H. Liu, ``Force-based algorithm for
  motion planning of large agent,'' \emph{IEEE Transactions on Cybernetics},
  2020.

\bibitem{grover2020deadlock}
J.~S. Grover, C.~Liu, and K.~Sycara, ``Deadlock analysis and resolution for
  multi-robot systems,'' in \emph{International Workshop on the Algorithmic
  Foundations of Robotics}.\hskip 1em plus 0.5em minus 0.4em\relax Springer,
  2020, pp. 294--312.

\bibitem{dergachev2021distributed}
S.~Dergachev and K.~Yakovlev, ``Distributed multi-agent navigation based on
  reciprocal collision avoidance and locally confined multi-agent path
  finding,'' in \emph{2021 IEEE 17th International Conference on Automation
  Science and Engineering (CASE)}.\hskip 1em plus 0.5em minus 0.4em\relax IEEE,
  2021, pp. 1489--1494.

\bibitem{hou2022enhanced}
J.~Hou, X.~Zhou, Z.~Gan, and F.~Gao, ``Enhanced decentralized autonomous aerial
  robot teams with group planning,'' \emph{IEEE Robotics and Automation
  Letters}, vol.~7, no.~4, pp. 9240--9247, 2022.

\bibitem{farouki2012bernstein}
R.~T. Farouki, ``The bernstein polynomial basis: A centennial retrospective,''
  \emph{Computer Aided Geometric Design}, vol.~29, no.~6, pp. 379--419, 2012.

\bibitem{mellinger2012mixed}
D.~Mellinger, A.~Kushleyev, and V.~Kumar, ``Mixed-integer quadratic program
  trajectory generation for heterogeneous quadrotor teams,'' in \emph{Robotics
  and Automation (ICRA), 2012 IEEE International Conference on}.\hskip 1em plus
  0.5em minus 0.4em\relax IEEE, 2012, pp. 477--483.

\bibitem{okumura2022priority}
K.~Okumura, M.~Machida, X.~D{\'e}fago, and Y.~Tamura, ``Priority inheritance
  with backtracking for iterative multi-agent path finding,'' \emph{Artificial
  Intelligence}, vol. 310, p. 103752, 2022.

\bibitem{boyd2004convex}
S.~Boyd, S.~P. Boyd, and L.~Vandenberghe, \emph{Convex optimization}.\hskip 1em
  plus 0.5em minus 0.4em\relax Cambridge university press, 2004.

\bibitem{okumura2021iterative}
K.~Okumura, Y.~Tamura, and X.~Défago, ``Iterative refinement for real-time
  multi-robot path planning,'' in \emph{2021 IEEE/RSJ International Conference
  on Intelligent Robots and Systems (IROS)}, 2021, pp. 9690--9697.

\bibitem{hornung2013octomap}
A.~Hornung, K.~M. Wurm, M.~Bennewitz, C.~Stachniss, and W.~Burgard, ``Octomap:
  An efficient probabilistic 3d mapping framework based on octrees,''
  \emph{Autonomous robots}, vol.~34, no.~3, pp. 189--206, 2013.

\bibitem{foltin2011automated}
M.~Foltin, ``Automated maze generation and human interaction,'' \emph{Brno:
  Masaryk University Faculty Of Informatics}, 2011.

\bibitem{cplex201612}
I.~CPLEX, ``12.7. 0 user’s manual,'' 2016.

\bibitem{preiss2017crazyswarm}
J.~A. Preiss, W.~Honig, G.~S. Sukhatme, and N.~Ayanian, ``Crazyswarm: A large
  nano-quadcopter swarm,'' in \emph{International Conference on Robotics and
  Automation (ICRA)}.\hskip 1em plus 0.5em minus 0.4em\relax IEEE, 2017, pp.
  3299--3304.

\end{thebibliography}

\end{document}